\documentclass[lettersize,journal]{IEEEtran}
\usepackage{amsmath,amsfonts}
\usepackage{algorithmic}
\usepackage{algorithm}
\usepackage{array}
\usepackage[caption=false,font=normalsize,labelfont=sf,textfont=sf]{subfig}
\usepackage{textcomp}
\usepackage{stfloats}
\usepackage{url}
\usepackage{verbatim}
\usepackage{graphicx}
\usepackage{cite}
\begin{document}

\title{Data-driven mode shape selection and model-based vibration suppression of 3-\underline{R}RR parallel manipulator with flexible actuation links}

\author{Dingxu Guo, Jian Xu, Shu Zhang
\thanks{D. Guo, J. Xu and S. Zhang are with the School of Aerospace Engineering
and Applied Mechanics, Tongji University, Shanghai, China.(Corresponding
author e-mail: zhangshu@tongji.edu.cn)}
\thanks{}}

\markboth{}%
{Shell \MakeLowercase{\textit{et al.}}: A Sample Article Using IEEEtran.cls for IEEE Journals}

\IEEEpubid{}

\maketitle

\begin{abstract}
The mode shape function is difficult to determine in modeling manipulators with flexible links using the assumed mode method. In this paper, for a planar 3-RRR parallel manipulator with flexible actuation links, we provide a data-driven method to identify the mode shape of the flexible links and propose a model-based controller for the vibration suppression. By deriving the inverse kinematics of the studied mechanism in analytical form, the dynamic model is established by using the assumed mode method. To select the mode shape function, the software of multi-body system dynamics is used to simulate the dynamic behavior of the mechanism, and then the data-driven method which combines the DMD and SINDy algorithms is employed to identify the reasonable mode shape functions for the flexible links. To suppress the vibration of the flexible links, a state observer for the end-effector is constructed by a neural network, and the model-based control law is designed on this basis. In comparison with the model-free controller, the proposed controller with developed dynamic model has promising performance in terms of tracking accuracy and vibration suppression.
\end{abstract}


\section{Introduction}
\IEEEPARstart{A}{s} the industry evolves, current industrial manipulators require more precision and higher speed to improve product quality and reduce time consumption\cite{ref1}. To guarantee the accuracy, typical industrial serial manipulators have large mass to increase the stiffness of the structure. However, the large mass also results in a significant decrease in the mobile capability of the manipulators due to the output torque limitations of actuators \cite{ref2}. Owing to the advantages of high operation speed, low energy consumption, and high compliance, manipulators with lightweight links are increasingly applied in many areas such as rapid sorting and assembly\cite{ref3,ref4}. However, the lightweight design always leads to the decrease in structural stiffness, and the elastic vibration of the lightweight links during high-speed manipulation negatively affects the accuracy and quality of operation of the manipulators\cite{ref5,ref6}. Maintaining the lightweight design while being able to reduce the impact of vibration is a challenging issue for the manipulators\cite{ref7}.

Due to the coupling of different branch chains, parallel manipulators have higher stiffness to weight ratios and higher accuracy compared with serial manipulators\cite{ref8}. Meanwhile, since the motors of typical parallel manipulators are usually fixed on the base, the links of parallel manipulators will not carry the mass and inertia of the motors, which makes the parallel manipulator easier to realize the lightweight of the links while ensuring the accuracy. However, the nonlinear geometric constraints of the parallel mechanism and the deformation generated by the lightweight links result in the complex dynamics behavior of the parallel manipulators with flexible links\cite{ref9}. For instance, in the case of the planar 3-RRR parallel manipulator with flexible links, Zhang et al.\cite{ref10} experimentally reveal that the residual vibration after the high-speed movement of the mechanism generates self-excited vibration and affects the positioning accuracy, indicating the flexible links in parallel manipulators present a challenge for the dynamic modeling and control strategy design.

Parallel manipulators with flexible links are modeled as a continuum system with an infinite number of degrees of freedom and the dynamic models are characterized by hybrid ordinary differential equations-partial differential equations (ODEs-PDEs)\cite{ref11,ref12}. For the above system, an analytical solution is difficult to obtain, so it is necessary to simplify. For this purpose, a certain flexible link is treated as an Euler-Bernoulli beam or a Timuchenko beam and the continuum beam can be reduced to finite degrees of freedom by using assumed mode method (AMM)\cite{ref13}, finite element method (FEM)\cite{ref14}, or lumped parameter method\cite{ref15}. Compared with the other methods, the AMM is characterized by a small amount of computation and convenience in control strategy design when the boundary conditions of the flexible links are determined\cite{ref16}, and thus it is widely used in model-based control of parallel manipulators with flexible links.

In the AMM, the continuous displacement of flexible links is discretized into the truncated finite modal series that are composed of mode shape functions and time-varying mode amplitudes\cite{ref17}. Wherein, the selection of the mode shape function is closely related to the configuration of the mechanism, and different mode shape functions have an intrinsic influence on the AMM. Ebrahimi S et al.\cite{ref18} analyzed the dynamic characteristics of a 3-RPR rigid-flexible coupling parallel manipulator by using AMM and simultaneously chose two possible mode shape functions, i.e., fixed-pinned and fixed-free, to establish different models and compared them. For the two different configurations of 3-PRR parallel manipulators with flexible intermediate links,\cite{ref19} and\cite{ref20} give two different modal shape functions of the AMM through experiments. Therefore, the selection of the mode shape function is a primary issue for the AMM. The identification of the modal shape function of the mechanism through experimental data acquisition is a reliable method, but it involves high costs and cannot be extended to other mechanisms. As a powerful complement, the simulation of the mechanism by commercial software and then the identification of the modal shape function using the collected data is an economical and general method\cite{ref21}.

The deformation of the flexible links not only induces complex dynamic behaviors of the parallel manipulator but also decreases the tracking or positioning accuracy of the end-effector\cite{ref22}. Therefore, it is critical to design an effective controller to suppress the vibration of the flexible links. For serial flexible manipulators, many control methods for vibration suppression are widely used, including input shaping control\cite{ref23}, singular perturbation control\cite{ref24}, optimal control\cite{ref25}, active vibration control with vibration actuators\cite{ref26}, etc. However, due to the structural complexity, the controller design for both trajectory tracking and vibration suppression of the parallel manipulator with flexible links is still a challenging issue. The parallel mechanism is characterized by the different branch chains being coupled via the end-effector. In the case of the parallel manipulator with flexible links, the vibration of the flexible links tends to cause the jittering of the end-effector. Therefore, to a certain extent, the overall state of the parallel manipulator with flexible links can be reflected by the state of the end-effector. However, due to the geometric nonlinearity of the parallel mechanism and the deformation generated by the flexible links, it is difficult to obtain the state of the end-effector in an analytic form by means of the feedback signal of the motors and deformation sensors. In recent years, artificial neural networks have been widely applied to the field of control owing to their excellent approximation capability\cite{ref27,ref28}. Model-based control combined with the artificial neural network is a subject worthy of further study in the field of vibration suppression.

This paper aims at providing a promising dynamic model for the parallel manipulator with flexible links and designing an efficient model-based control for vibration suppression. Compared with previous studies, the main contributions of this paper are as follows:

1)	The inverse kinematics with an analytic form of a planar 3RRR parallel manipulator with flexible actuation links is derived. Using the AMM to discretize the flexible links, the dynamic model of the aforementioned mechanism is established by the Lagrangian method.

2)	A data-driven method of the mode shape function identification is proposed. Firstly, the deformation data of the flexible links is generated by the numerical simulation. Then, the data-driven method is employed to find the reasonable mode shape function for the flexible links.

3)	A state observer for the end-effector is constructed by a neural network, and a model-based controller is designed on this basis. The proposed controller can effectively suppress the vibration of the mechanism and is suitable for the cases in which the feedback frequencies of the flexible sensors are limited.

The remainder of this paper is organized as follows. In Section 2, the structure and the selection of rigidity and flexibility of the planar 3RRR parallel manipulator are illustrated, and the inverse kinematics with an analytic form of the mechanism is derived. In Section 3, the dynamic model of the mechanism is established via the AMM, and the boundary conditions of the flexible links remain to be determined. Section 4 introduces the data-driven method to identify the mode shape function. The construction of the state observer for the end-effector and the design of the model-based controller is illustrated in Section 5. In Section 6, the case investigations and results discussion are presented. The conclusion of this paper is described in Section 7.

\section{Structure description and kinematic analysis}

\begin{figure*}[!t]
\centering
\includegraphics[width=5in]{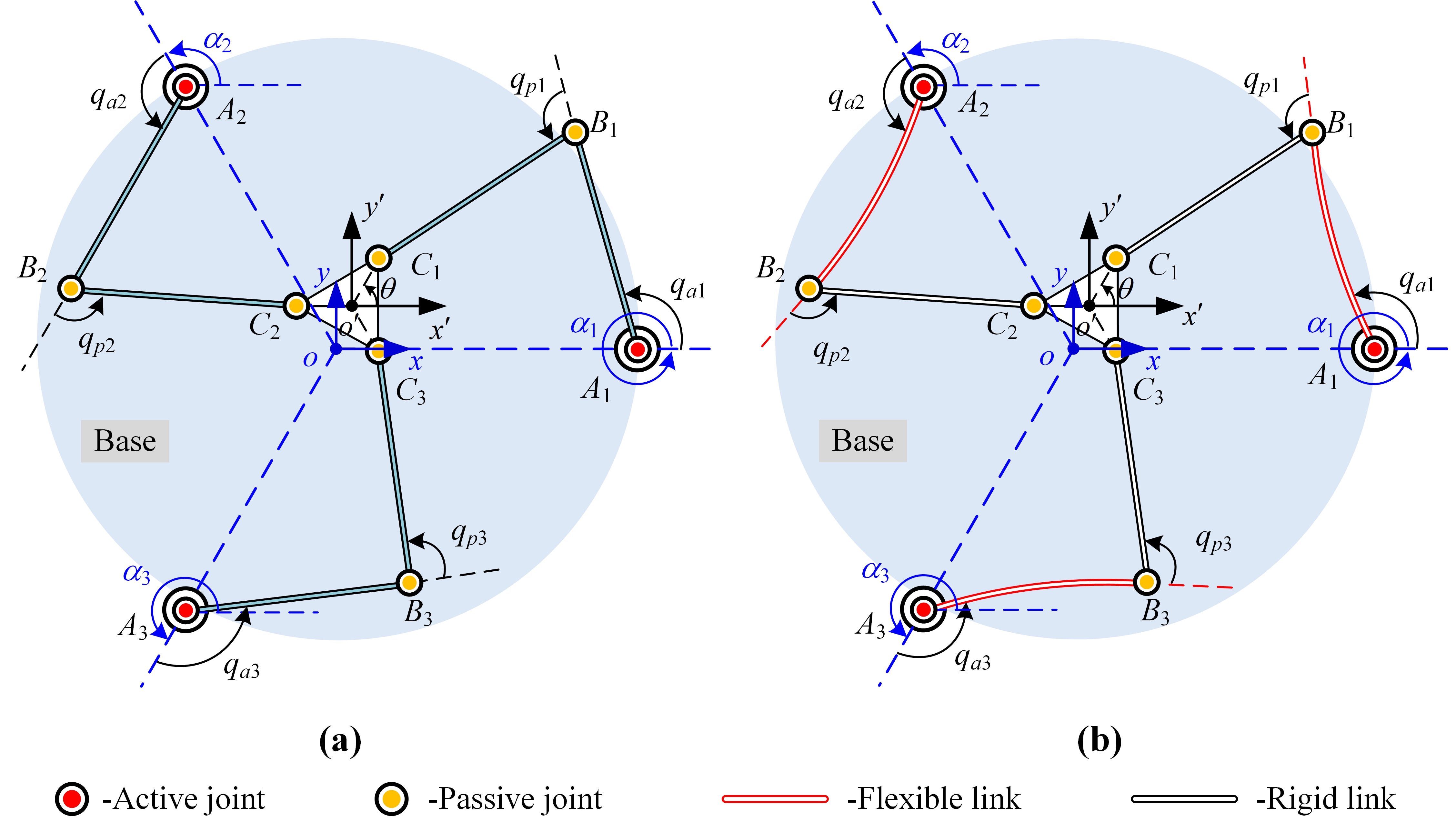}
\caption{(a) The structure diagram of the planar 3-RRR parallel manipulator; (b) the motion schematic diagram of the planar 3-RRR parallel manipulator with flexible actuation links.}
\label{fig_1}
\end{figure*}

\subsection{Structure description}
The structure of the planar 3-\underline{R}RR parallel manipulator (PM) is shown in Fig. \ref{fig_1}(a), which consists of a fixed base, three identical \underline{R}RR serial branches, and a moving platform. Each actuation revolute joint is driven at $A_i$ by a motor, and other revolute joints, i.e., $B_i$ and $C_i$, are passive, $i=1,2,3$. Each branch is composed of an actuation link $A_iB_i$ and an intermediate link $B_iC_i$ of lengths of $l_1$ and $l_2$, respectively. The moving platform is connected to intermediate links at the distal end, i.e., $C_i$, $i=1,2,3$. Denote the angles of the actuation links and the intermediate links by $q_{ai}$ and $q_{pi}$, $i=1,2,3$. The fixed base with a radius $R$ is bisected by $A_1$, $A_2$, $A_3$. The moving platform, i.e., the end-effector, is of a round plate ($C_1C_2C_3$) with a radius of $r$. 

In the previous studies of the rigid-flexible coupling 3-RRR PM, the actuation links always had higher stiffness than the intermediate links, and thus only the intermediate links were assumed to be flexible. In\cite{ref24} and\cite{ref29}, the pinned–pinned boundary condition was applied to the intermediate links, implying that the deformation of the intermediate links barely affected the end-effector. To further pursue the lightweight design, it is necessary to investigate the dynamic behavior of the 3-\underline{R}RR PM with all the links lightweighted. To this end, we use the software called automated dynamic analysis of mechanical systems (ADAMS) to simulate the 3-RRR PM with all the links lightweighted. The pre-simulation results show that, when the parameters of the actuation links and the intermediate links are consistent, the deformation of the actuation link in each branch is much larger than that of the intermediate link. Meanwhile, the deformation of the actuation links has a great impact on the state of the end-effector. In contrast, since no active torques are applied to the intermediate links, the deformation of the intermediate links is too small to be negligible. Therefore, for the 3-\underline{R}RR PM with all the lightweighted links, the actuation links are considered flexible and the intermediate links are considered rigid in this paper, and the motion schematic diagram of the mechanism is shown in Fig. \ref{fig_1}(b). The above pre-simulation results will be shown in the Appendix A.

\subsection{Inverse kinematics and Jacobian matrix}

\begin{figure}[!t]
\centering
\includegraphics[width=3in]{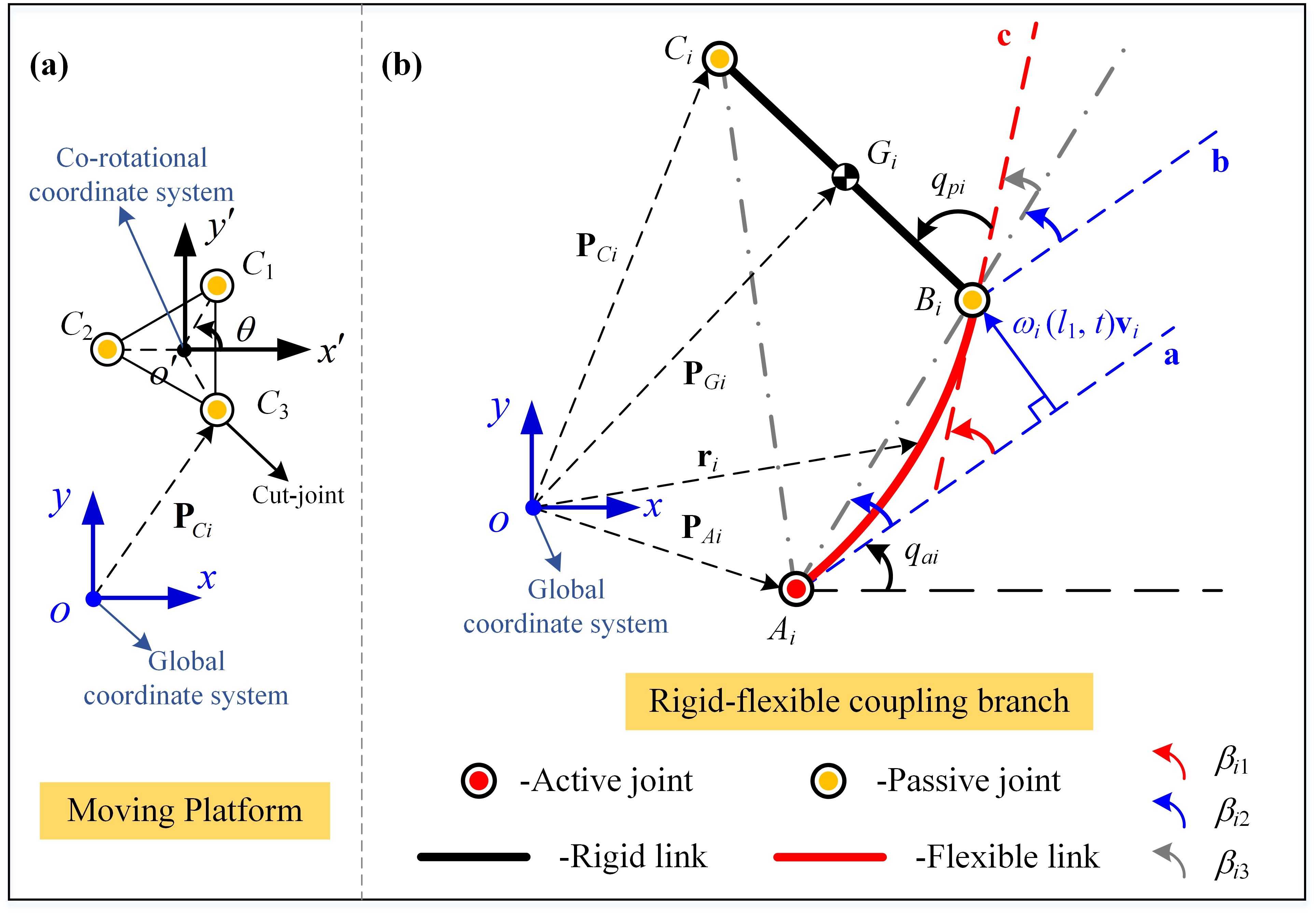}
\caption{Geometric model of 3-RRR PMFAL for (a) the moving platform and (b) the rigid-flexible coupling branch.}
\label{fig_2}
\end{figure}

Due to the requirements of dynamic modeling and trajectory planning, it is necessary to formulate the inverse kinematics of the 3\underline{R}RR parallel manipulator with flexible actuation links (PMFAL). The objective of the inverse kinematics is to obtain   and   from the state of the end-effector and the deformation of the actuation link. As shown in Fig. \ref{fig_1}(b), the coordinates of the 3-\underline{R}RR PMFAL are defined as follows: the global coordinate system $\Re:o - xyz$ which is mounted on the center of the fixed base, and the co-rotational coordinate system $\Re':o'-x'y'z'$ which is mounted on the center of the moving platform. As shown in Fig. 2(a), denote by ${{\mathbf{q}}_{\text{e}}} = {\left[ {\begin{array}{*{20}{c}}
  x&y&\theta  
\end{array}} \right]^{\text{T}}}$ the position and angle of the moving platform in $\Re$, the position of ${C_i}\left( {i = 1,2,3} \right)$ can be obtained as
\begin{equation}
\label{eq_1}
{{\mathbf{P}}_{Ci}} = {\left[ {\begin{array}{*{20}{c}}
  {x + r\cos \left( {\theta  + {\alpha _i}} \right)}&{y + r\sin \left( {\theta  + {\alpha _i}} \right)} 
\end{array}} \right]^{\text{T}}},
\end{equation}
where ${\alpha _i} = {{\left( {i - 1} \right)2\pi } \mathord{\left/
 {\vphantom {{\left( {i - 1} \right)2\pi } 3}} \right.
 \kern-\nulldelimiterspace} 3},{\text{ }}i = 1,2,3$. The flexible actuation links are modeled as Euler-Bernoulli beams in this paper. The configuration of a branch with rigid-flexible coupling is shown in Fig. \ref{fig_2}(b), where the deformation of $A_iB_i$ is a variable about time and space and can be expressed as
\begin{equation}
\label{eq_2}
{\omega _i}\left( {x,t} \right),{\text{ }}x \in \left[ {0,{\text{ }}l} \right],{\text{ }}i = 1,2,3.
\end{equation}
Herein, the AMM is used to model the deformation of the flexible links. Utilizing the AMM, ${\omega _i}\left( {x,t} \right)$ can be obtained as
\begin{equation}
\label{eq_3}
{\omega _i}\left( {x,t} \right) = \sum\limits_{j = 1}^\infty  {{\phi _{ij}}\left( x \right)} {q_f}_{ij}\left( t \right),{\text{ }}i = 1,2,3,
\end{equation}
where ${\phi _{ij}}\left( x \right)$ represents the ${j_{th}}$ order normalized mode shape function of the actuation link $A_iB_i$, ${q_f}_{ij}\left( t \right)$ represents the time-dependent mode coordinate corresponding to ${\phi _{ij}}\left( x \right)$. To facilitate computation, the modal truncation method is employed, and ${\omega _i}\left( {x,t} \right)$ can be rewritten as
\begin{equation}
\label{eq_4}
{\omega _i}\left( {x,t} \right) \approx \sum\limits_{j = 1}^n {{\phi _{ij}}\left( x \right)} {q_f}_{ij}\left( t \right),{\text{ }}i = 1,2,3,
\end{equation}
where $n$ denotes the number of selected modes. In $\Re:o - xyz$, the position vector of any point ($x$) on the link $A_iB_i$, i.e., ${{\mathbf{r}}_i}$ in Fig. \ref{fig_2}(b), can be expressed as
\begin{equation}
\label{eq_5}
{{\mathbf{r}}_i} = {{\mathbf{P}}_{Ai}} + x{{\mathbf{u}}_i} + {\omega _i}\left( {x,t} \right){{\mathbf{v}}_i},
\end{equation}
where ${{\mathbf{P}}_{Ai}} = {\left[ {\begin{array}{*{20}{c}}
  {R\cos {\alpha _i}}&{R\sin {\alpha _i}} 
\end{array}} \right]^{\text{T}}}$, and $\mathbf{u}_i$, $\mathbf{v}_i$ represent respectively the unit vector in the direction of $A_iB_i$ in its undeformed status (line $\mathbf{a}$ in Fig. 2(b)) and the direction of the deformation of link $A_iB_i$, namely, ${{\mathbf{u}}_i} = {\left[ {\begin{array}{*{20}{c}}
  {\cos \left( {{\alpha _i} + {q_{ai}}} \right)}&{\sin \left( {{\alpha _i} + {q_{ai}}} \right)} 
\end{array}} \right]^{\text{T}}}$, ${{\mathbf{v}}_i} = {\left[ {\begin{array}{*{20}{c}}
  { - \sin \left( {{\alpha _i} + {q_{ai}}} \right)}&{\cos \left( {{\alpha _i} + {q_{ai}}} \right)} 
\end{array}} \right]^{\text{T}}}$. The position vector of the center of mass (COM) of link $B_iC_i$, i.e., ${{\mathbf{P}}_{Gi}}$ in Fig. \ref{fig_2}(b), is given by
\begin{equation}
\label{eq_6}
{{\mathbf{P}}_{Gi}} = {{\mathbf{P}}_{Ai}} + {l_1}{{\mathbf{u}}_i} + {\omega _i}\left( {{l_1},t} \right){{\mathbf{v}}_i} + {l_c}{{\mathbf{w}}_i},
\end{equation}
where ${l_c}$ represents the distance between the COM and the initial point of link $B_iC_i$ and ${{\mathbf{w}}_i}$ represents the unit vector in the direction of $B_iC_i$. The expression of ${{\mathbf{w}}_i}$ can be written as
\begin{equation}
\label{eq_7}
{{\mathbf{w}}_i} = {\left[ {\begin{array}{*{20}{c}}
  {\cos \left( {{\alpha _i} + {q_{ai}} + {\beta _{i1}} + {q_{pi}}} \right)}&{\sin \left( {{\alpha _i} + {q_{ai}} + {\beta _{i1}} + {q_{pi}}} \right)} 
\end{array}} \right]^{\text{T}}},
\end{equation}
where
\begin{equation}
\label{eq_8}
{\beta _{i1}} = {\left. {\frac{{\partial {\omega _i}\left( {x,t} \right)}}{{\partial x}}} \right|_{x = {l_1}}}.
\end{equation}
As shown in Fig. \ref{fig_2}(b), $\mathbf{a}$ and $\mathbf{c}$ are the tangents of flexible link $A_iB_i$ at points $A_i$ and $B_i$, respectively, and $\mathbf{b}$ is an auxiliary line parallel to line $\mathbf{a}$. Using the well known trigonometric equality, one obtains
\begin{equation}
\label{eq_9}
\angle {A_i}{B_i}{C_i} = {\text{arccos}}\left( {\frac{{{{\left| {{{\mathbf{A}}_i}{{\mathbf{B}}_i}} \right|}^2} + {{\left| {{{\mathbf{B}}_i}{{\mathbf{C}}_i}} \right|}^2} - {{\left| {{{\mathbf{A}}_i}{{\mathbf{C}}_i}} \right|}^2}}}{{\left( {2\left| {{{\mathbf{A}}_i}{{\mathbf{C}}_i}} \right|} \right)}}} \right),
\end{equation}
${\beta _{i2}}$ and ${\beta _{i3}}$ marked in Fig. \ref{fig_2}(b) can be obtained as
\begin{equation}
\label{eq_10}
{\beta _{i2}} = \arctan \left( {{{{\omega _i}\left( {{l_1},t} \right)} \mathord{\left/
 {\vphantom {{{\omega _i}\left( {{l_1},t} \right)} {{l_1}}}} \right.
 \kern-\nulldelimiterspace} {{l_1}}}} \right),
\end{equation}
\begin{equation}
\label{eq_11}
{\beta _{i3}} = {\beta _{i2}} - {\beta _{i1}}.
\end{equation}
Finally, ${q_{pi}}$ is given by
\begin{equation}
\label{eq_12}
{q_{pi}} = \pi  - \angle {A_i}{B_i}{C_i} - {\beta _{i3}},{\text{ }}i = 1,2,3.
\end{equation}
According to Fig. \ref{fig_2}(b), the position vector of ${{\mathbf{C}}_i}$ can also be expressed as
\begin{equation}
\label{eq_13}
{{\mathbf{P}}_{Ci}} = {{\mathbf{P}}_{Ai}} + {l_1}{{\mathbf{u}}_i} + {\omega _i}\left( {{l_1},t} \right){{\mathbf{v}}_i} + {l_2}{{\mathbf{w}}_i}.
\end{equation}
Combining Eqs. (\ref{eq_1}) and (\ref{eq_13}) yields
\begin{equation}
\label{eq_14}
\begin{gathered}
  {\left[ {\begin{array}{*{20}{c}}
  {x + r\cos \left( {\theta  + {\alpha _i}} \right)}&{y + r\sin \left( {\theta  + {\alpha _i}} \right)} 
\end{array}} \right]^{\text{T}}} \hfill \\
   = {{\mathbf{P}}_{Ai}} + {l_1}{{\mathbf{u}}_i} + {\omega _i}\left( {{l_1},t} \right){{\mathbf{v}}_i} + {l_2}{{\mathbf{w}}_i}. \hfill \\ 
\end{gathered} 
\end{equation}
Based on Eq. \ref{eq_14}, $q_{ai}$ can be solved as
\begin{equation}
\label{eq_15}
{q_{ai}} = \left\{ {\begin{array}{*{20}{c}}
  {\pi  - \arcsin \left( {\frac{{bc - ad}}{{{c^2} + {d^2}}}} \right) - {\alpha _i},{\text{ }}i = 1,2,} \\ 
  {2\pi  + \arcsin \left( {\frac{{bc - ad}}{{{c^2} + {d^2}}}} \right) - {\alpha _i},{\text{ }}i = 3.} 
\end{array}} \right.
\end{equation}
where

$a = x + r\cos \left( {\theta  + {\alpha _i}} \right) - R\cos {\alpha _i}$,

$b = y + r\sin \left( {\theta  + {\alpha _i}} \right) - R\sin {\alpha _i}$,

$c = {l_1} + {l_2}\cos \left( {{\beta _{i1}} + {q_{pi}}} \right)$,

$d = {\omega _i}\left( {{l_1},t} \right) + {l_2}\sin \left( {{\beta _{i1}} + {q_{pi}}} \right)$.

In the next, we will carry out the velocity analysis that is crucial for the formulation of the dynamic model. Differentiating both sides of Eq. (\ref{eq_14}) with respect to time yield
\begin{equation}
\label{eq_16}
\begin{gathered}
  \left[ {\begin{array}{*{20}{c}}
  {\dot x + r\sin \left( {\theta  + {\alpha _i}} \right)\dot \theta } \\ 
  {\dot y + r\cos \left( {\theta  + {\alpha _i}} \right)\dot \theta } 
\end{array}} \right] = l{{\mathbf{\omega }}_{i1}} \times {{\mathbf{u}}_i} + {\left. {\frac{{\partial {\omega _i}}}{{\partial t}}} \right|_{x = {l_1}}}{{\mathbf{v}}_i} \\ 
   + {\omega _i}\left( {{l_1},t} \right){{\mathbf{\Omega }}_{i1}} \times {{\mathbf{v}}_i} + l{{\mathbf{\Omega }}_{i2}} \times {{\mathbf{w}}_i}, \\ 
\end{gathered} 
\end{equation}
where ${{\mathbf{\Omega }}_{i1}}$, ${{\mathbf{\Omega }}_{i2}}$ represent respectively the angular velocities of link $A_iB_i$ and link $B_iC_i$, and ${{\mathbf{\Omega }}_{i1}} \times {{\mathbf{u}}_i} = {\dot q_{ai}}{{\mathbf{v}}_i}$, ${{\mathbf{\Omega }}_{i1}} \times {{\mathbf{v}}_i} =  - {\dot q_{ai}}{{\mathbf{u}}_i}$. Eq. (\ref{eq_16}) can be simplified as
\begin{equation}
\label{eq_17}
\begin{gathered}
  {{{\mathbf{\dot P}}}_{Ci}} = \left[ {l{{\dot q}_{ai}} + {{\left. {{{\partial {\omega _i}\left( {x,t} \right)} \mathord{\left/
 {\vphantom {{\partial {\omega _i}\left( {x,t} \right)} {\partial t}}} \right.
 \kern-\nulldelimiterspace} {\partial t}}} \right|}_{x = {l_1}}}} \right]{{\mathbf{v}}_i} \\ 
   - {\omega _i}\left( {{l_1},t} \right){{\dot q}_{ai}}{{\mathbf{u}}_i} + l{{\mathbf{\Omega }}_{i2}} \times {{\mathbf{w}}_i}. \\ 
\end{gathered}
\end{equation}
Based on Eq. (\ref{eq_17}), a direct calculation shows that
\begin{equation}
\label{eq_18}
\begin{gathered}
  {{{\mathbf{\dot P}}}_{Ci}} \cdot {{\mathbf{w}}_i} = \left[ {{l_1}{{\dot q}_{ai}} + {{\left. {{{\partial {\omega _i}\left( {x,t} \right)} \mathord{\left/
 {\vphantom {{\partial {\omega _i}\left( {x,t} \right)} {\partial t}}} \right.
 \kern-\nulldelimiterspace} {\partial t}}} \right|}_{x = {l_1}}}} \right]{{\mathbf{v}}_i} \cdot {{\mathbf{w}}_i} \hfill \\
   - {\omega _i}\left( {{l_1},t} \right){{\dot q}_{ai}}{{\mathbf{u}}_i} \cdot {{\mathbf{w}}_i} + {l_2}{{\mathbf{\Omega }}_{i2}} \times {{\mathbf{w}}_i} \cdot {{\mathbf{w}}_i}, \hfill \\
  {{\dot q}_{ai}} = \frac{{{{{\mathbf{\dot P}}}_{Ci}} \cdot {{\mathbf{w}}_i} - {{\left. {{{\partial {\omega _i}\left( {x,t} \right)} \mathord{\left/
 {\vphantom {{\partial {\omega _i}\left( {x,t} \right)} {\partial t}}} \right.
 \kern-\nulldelimiterspace} {\partial t}}} \right|}_{x = {l_1}}}{{\mathbf{v}}_i} \cdot {{\mathbf{w}}_i}}}{{{l_1}{{\mathbf{v}}_i} \cdot {{\mathbf{w}}_i} - {\omega _i}\left( {{l_1},t} \right){{\mathbf{u}}_i} \cdot {{\mathbf{w}}_i}}}. \hfill \\ 
\end{gathered} 
\end{equation}
and
\begin{equation}
\label{eq_19}
\begin{gathered}
  {{{\mathbf{\dot P}}}_{Ci}} \times {{\mathbf{w}}_i} = \left[ {{l_1}{{\dot q}_{ai}} + {{\left. {{{\partial {\omega _i}\left( {x,t} \right)} \mathord{\left/
 {\vphantom {{\partial {\omega _i}\left( {x,t} \right)} {\partial t}}} \right.
 \kern-\nulldelimiterspace} {\partial t}}} \right|}_{x = {l_1}}}} \right]{{\mathbf{v}}_i} \times {{\mathbf{w}}_i} \hfill \\
   - {\omega _i}\left( {{l_1},t} \right){{\dot q}_{ai}}{{\mathbf{u}}_i} \times {{\mathbf{w}}_i} + {l_2}{{\mathbf{\Omega }}_{i2}} \times {{\mathbf{w}}_i} \times {{\mathbf{w}}_i}, \hfill \\
  {{\mathbf{\Omega }}_{i2}} = \frac{{\left( {{{{\mathbf{\dot P}}}_{Ci}} - \left( {{l_1}{{\dot q}_{ai}} + {{\left. {\frac{{\partial {\omega _i}}}{{\partial t}}} \right|}_{x = {l_1}}}} \right){{\mathbf{v}}_i} - {\omega _i}\left( {{l_1},t} \right){{\dot q}_{ai}}{{\mathbf{u}}_i}} \right) \times {{\mathbf{w}}_i}}}{{{l_2}}}. \hfill \\ 
\end{gathered}
\end{equation}
As shown in Eq. (\ref{eq_4}), the continuum beam can be reduced to finite degrees of freedom by using the modal truncation method. Denote the generalized coordinates of the 3-\underline{R}RR PMFAL by
\begin{equation}
\label{eq_20}
{\mathbf{q}} = {\left[ {\begin{array}{*{20}{c}}
  {{{\mathbf{q}}_{\text{e}}}^{\text{T}}}&{{\mathbf{q}}_{\text{f}}^{\text{T}}} 
\end{array}} \right]^{\text{T}}},
\end{equation}
where ${{\mathbf{q}}_{\text{f}}}$ is the vector of the mode coordinates and ${{\mathbf{q}}_{\text{f}}} = {\left[ {\begin{array}{*{20}{c}}
  {{q_f}_{11}}& \cdots &{\begin{array}{*{20}{c}}
  {{q_f}_{1n}}&{{q_f}_{21}}&{\begin{array}{*{20}{c}}
   \cdots &{{q_f}_{2n}}&{\begin{array}{*{20}{c}}
  {{q_f}_{31}}& \cdots &{{q_f}_{3n}} 
\end{array}} 
\end{array}} 
\end{array}} 
\end{array}} \right]^{\text{T}}}$. Meanwhile, we define
\begin{equation}
\label{eq_21}
\begin{gathered}
  {{\mathbf{q}}_{\text{d}}} = {\left[ {\begin{array}{*{20}{c}}
  {{\mathbf{q}}_{\text{a}}^{\text{T}}}&{{\mathbf{q}}_{\text{f}}^{\text{T}}} 
\end{array}} \right]^{\text{T}}},{\text{ }} \hfill \\
  {{\mathbf{q}}_{\text{w}}} = {\left[ {\begin{array}{*{20}{c}}
  {{\mathbf{q}}_{\text{a}}^{\text{T}}}&{{\mathbf{q}}_{\text{p}}^{\text{T}}}&{{\mathbf{q}}_{\text{f}}^{\text{T}}} 
\end{array}} \right]^{\text{T}}}, \hfill \\ 
\end{gathered} 
\end{equation}
where ${{\mathbf{q}}_{\text{a}}} = {\left[ {\begin{array}{*{20}{c}}
  {{q_{a1}}}&{{q_{a2}}}&{{q_{a3}}} 
\end{array}} \right]^{\text{T}}}$ and ${{\mathbf{q}}_{\text{p}}} = {\left[ {\begin{array}{*{20}{c}}
  {{q_{p1}}}&{{q_{p2}}}&{{q_{p3}}} 
\end{array}} \right]^{\text{T}}}$. Based on Eqs. (\ref{eq_12}) and (\ref{eq_15}), two Jacobian matrices are given by
\begin{equation}
\label{eq_22}
\begin{gathered}
  {{{\mathbf{\dot q}}}_{\text{d}}} = {\mathbf{J\dot q}}, \hfill \\
  {{{\mathbf{\dot q}}}_{\text{w}}} = {\mathbf{S\dot q}}, \hfill \\ 
\end{gathered} 
\end{equation}
where 

${\mathbf{J}} = \left[ {\begin{array}{*{20}{c}}
  {\frac{{\partial {{\mathbf{q}}_{\text{a}}}}}{{\partial {\mathbf{q}}_{\text{e}}^{\text{T}}}}}&{\frac{{\partial {{\mathbf{q}}_{\text{a}}}}}{{\partial {\mathbf{q}}_{\text{f}}^{\text{T}}}}} \\ 
  {{{\mathbf{0}}_{3n \times 3}}}&{{{\mathbf{I}}_{3n \times 3n}}} 
\end{array}} \right] = \left[ {\begin{array}{*{20}{c}}
  {{{\mathbf{J}}_{{\text{ax}}}}}&{{{\mathbf{J}}_{{\text{af}}}}} \\ 
  {{{\mathbf{0}}_{3 \times 3}}}&{{{\mathbf{I}}_{3n \times 3n}}} 
\end{array}}\right],$

${\mathbf{S}} = {\left[ {\begin{array}{*{20}{c}}
  {\frac{{\partial {\mathbf{q}}_{\text{a}}^{\text{T}}}}{{\partial {\mathbf{q}}}}}&{\frac{{\partial {\mathbf{q}}_{\text{p}}^{\text{T}}}}{{\partial {\mathbf{q}}}}}&{\begin{array}{*{20}{c}}
  {{{\mathbf{0}}_{3 \times 3n}}} \\ 
  {{{\mathbf{I}}_{3n \times 3n}}} 
\end{array}} 
\end{array}} \right]^{\text{T}}},$
\\ and ${\mathbf{I}}$ denotes the unit matrix and ${\mathbf{0}}$ is the zero matrix.

The inverse kinematics and the Jacobian matrices of 3-\underline{R}RR PMFAL provide necessary preparations for formulating the dynamic model by means of the Lagrangian method, as will be seen in the next section.

\section{Dynamic modeling}
The Lagrangian method is employed in this section to develop a dynamic model for the 3-\underline{R}RR PMFAL. To this end, the kinetic energy and potential energy are formulated as follows.

\subsection{Kinetic energy}
The movable parts of the 3-\underline{R}RR PMFAL consist of three types of components: the flexible actuation links, the rigid intermediate links, and the end-effector. The kinetic energy of flexible link $A_iB_i$ is expressed as
\begin{equation}
\label{eq_23}
{T_{i1}} = \frac{1}{2}\int_0^{{l_1}} {{\rho _i}{\mathbf{\dot r}}_i^{\text{T}}{{{\mathbf{\dot r}}}_i}} dx,{\text{ }}i = 1,2,3,
\end{equation}
where ${\rho _i}$ is the mass per unit length of $A_iB_i$, and the form of ${{\mathbf{r}}_i}$ can be found in Eq. (\ref{eq_5}). According to the Koenig Theorem, the kinetic energy of rigid link $B_iC_i$ can be written as
\begin{equation}
\label{eq_24}
{T_{i2}} = \frac{1}{2}{m_r}{\mathbf{\dot P}}_{Gi}^{\text{T}}{{\mathbf{\dot P}}_{Gi}} + \frac{1}{2}{J_r}{\mathbf{\Omega }}_{i2}^{\text{T}}{{\mathbf{\Omega }}_{i2}},{\text{ }}i = 1,2,3,
\end{equation}
where $m_r$ and $J_r$ represent respectively the mass and the rotational inertia about the COM of $B_iC_i$, and the form of ${{\mathbf{P}}_{Gi}}$ and ${{\mathbf{\Omega }}_{i2}}$ can be found in Eqs. (\ref{eq_6}) and (\ref{eq_19}) respectively. The kinetic energy of the end-effector is given by
\begin{equation}
\label{eq_25}
{T_3} = \frac{1}{2}{m_e}\left( {{{\dot x}^2} + {{\dot y}^2}} \right) + \frac{1}{2}{J_e}{\dot \theta ^2},
\end{equation}
where $m_e$ and $J_e$ represent the mass and the rotational inertia about the COM of the end-effector. Then the total kinetic energy of the 3-\underline{R}RR PMFAL can be written as
\begin{equation}
\label{eq_26}
T = \sum\limits_{i = 1}^3 {\sum\limits_{k = 1}^2 {{T_{ik}}} }  + {T_3}.
\end{equation}
\subsection{Potential energy}
Since the 3-\underline{R}RR PMFAL under consideration is assumed to be restricted to the horizontal plane, only elastic potential energy needs to be considered in this section. As the flexible actuation links are modeled as Euler-Bernoulli beams in this paper, the potential energy is caused by the bending deformation of link $A_iB_i$. The total potential energy is calculated as
\begin{equation}
\label{eq_27}
V = \frac{1}{2}\sum\limits_{i = 1}^3 {\int_0^{{l_1}} {{E_i}{I_i}{{\left( {{{{\partial ^2}{\omega _i}\left( {x,t} \right)} \mathord{\left/
 {\vphantom {{{\partial ^2}{\omega _i}\left( {x,t} \right)} {\partial {x^2}}}} \right.
 \kern-\nulldelimiterspace} {\partial {x^2}}}} \right)}^2}dx} } ,
\end{equation}
where $E_i$ is the Young’s modulus, and $I_i$ is the area moment of inertia of link $A_iB_i$.
\subsection{Lagrange equation}
The Lagrange equations of the 3-\underline{R}RR PMFAL are given by
\begin{equation}
\label{eq_28}
\frac{d}{{dt}}\left( {\frac{{\partial L}}{{\partial {{{\mathbf{\dot q}}}_{\text{w}}}}}} \right) - \frac{{\partial L}}{{\partial {{\mathbf{q}}_{\text{w}}}}} = {{\mathbf{Q}}_{\text{w}}},
\end{equation}
where $L=T-V$ and ${{\mathbf{Q}}_{\text{w}}}$ represents the external forces corresponding to ${{\mathbf{q}}_{\text{w}}}$. The resulting dynamic equation in the matrix form is given by
\begin{equation}
\label{eq_29}
{\mathbf{M}}\left( {{{\mathbf{q}}_{\text{w}}}} \right){{\mathbf{\ddot q}}_{\text{w}}} + {\mathbf{C}}\left( {{{\mathbf{q}}_{\text{w}}},{{{\mathbf{\dot q}}}_{\text{w}}}} \right){{\mathbf{\dot q}}_{\text{w}}} + {\mathbf{K}}\left( {{{\mathbf{q}}_{\text{w}}}} \right){{\mathbf{q}}_{\text{w}}} = {{\mathbf{Q}}_{\text{w}}},
\end{equation}
where ${\mathbf{M}}\left( {{{\mathbf{q}}_{\text{w}}}} \right)$, ${\mathbf{C}}\left( {{{\mathbf{q}}_{\text{w}}},{{{\mathbf{\dot q}}}_{\text{w}}}} \right)$ and ${\mathbf{K}}\left( {{{\mathbf{q}}_{\text{w}}}} \right)$ respectively represent the mass matrix, the Coriolis and centrifugal matrix and the stiffness matrix. The inputs of the 3-\underline{R}RR PMFAL come from the active joints or distributed smart materials such as the piezoelectric transducer (PZT). Therefore, the driving vector can be denoted by
\begin{equation}
\label{eq_30}
{\mathbf{\tau }} = {\left[ {\begin{array}{*{20}{c}}
  {{{\mathbf{\tau }}_{\text{a}}}^{\text{T}}}&{{{\mathbf{\tau }}_{\text{f}}}^{\text{T}}} 
\end{array}} \right]^{\text{T}}},
\end{equation}
where ${{\mathbf{\tau }}_{\text{a}}} = {\left[ {\begin{array}{*{20}{c}}
  {{\tau _{a1}}}&{{\tau _{a2}}}&{{\tau _{a3}}} 
\end{array}} \right]^{\text{T}}}$ and ${{\mathbf{\tau }}_{\text{f}}}$ is the vector of the input forces (or torques) corresponding to the mode coordinates vector ${{\mathbf{q}}_{\text{f}}}$. In addition, the generalized forces (or torques) corresponding to $\mathbf{q}$ can be denoted by $\mathbf{Q}$. According to the principle of virtual displacement\cite{ref30}, ${{\mathbf{Q}}_{\text{w}}}$, $\mathbf{Q}$ and ${\mathbf{\tau}}$ are related by
\begin{equation}
\label{eq_31}
{\mathbf{Q}} = {{\mathbf{J}}^{\text{T}}}{\mathbf{\tau }} = {{\mathbf{S}}^{\text{T}}}{{\mathbf{Q}}_{\text{w}}}.
\end{equation}
The proof of Eq. (\ref{eq_31}) can be found in Appendix B. Combining Eqs. (\ref{eq_29}) and (\ref{eq_31}), the dynamic equation of the 3-\underline{R}RR PMFAL can be rewritten as
\begin{equation}
\label{eq_32}
{\mathbf{\hat M\ddot q}} + {\mathbf{\hat C\dot q}} + {\mathbf{\hat Kq}} = {{\mathbf{J}}^{\text{T}}}{\mathbf{\tau}},
\end{equation}
where
\begin{equation*}
\begin{gathered}
  {\mathbf{\hat M}} = {{\mathbf{S}}^{\text{T}}}{\mathbf{MS}}, \hfill \\
  {\mathbf{\hat C}} = {{\mathbf{S}}^{\text{T}}}{\mathbf{M\dot S}} + {{\mathbf{S}}^{\text{T}}}{\mathbf{CS}}, \hfill \\
  {\mathbf{\hat K}} = {{\mathbf{S}}^{\text{T}}}{\mathbf{K}}. \hfill \\ 
\end{gathered}
\end{equation*}
Eq. (\ref{eq_32}) in partitioned form can be written as
\begin{equation}
\label{eq_33}
\begin{gathered}
  \left[ {\begin{array}{*{20}{c}}
  {{{{\mathbf{\hat M}}}_{{\text{rr}}}}}&{{{{\mathbf{\hat M}}}_{{\text{rf}}}}} \\ 
  {{\mathbf{\hat M}}_{{\text{rf}}}^{\text{T}}}&{{{{\mathbf{\hat M}}}_{{\text{ff}}}}} 
\end{array}} \right]\left[ {\begin{array}{*{20}{c}}
  {{\mathbf{\ddot T}}} \\ 
  {{{{\mathbf{\ddot q}}}_{\text{f}}}} 
\end{array}} \right] + \left[ {\begin{array}{*{20}{c}}
  {{{{\mathbf{\hat C}}}_{{\text{rr}}}}}&{{{{\mathbf{\hat C}}}_{{\text{rf}}}}} \\ 
  {{{{\mathbf{\hat C}}}_{{\text{fr}}}}}&{{{{\mathbf{\hat C}}}_{{\text{ff}}}}} 
\end{array}} \right]\left[ {\begin{array}{*{20}{c}}
  {{\mathbf{\dot T}}} \\ 
  {{{{\mathbf{\dot q}}}_{\text{f}}}} 
\end{array}} \right] \hfill \\
   + \left[ {\begin{array}{*{20}{c}}
  {{{\mathbf{0}}_{3 \times 3}}}&{{{\mathbf{0}}_{3 \times 3n}}} \\ 
  {{{\mathbf{0}}_{3n \times 3}}}&{{{{\mathbf{\hat K}}}_{{\text{ff}}}}} 
\end{array}} \right]\left[ {\begin{array}{*{20}{c}}
  {\mathbf{T}} \\ 
  {{{\mathbf{q}}_{\text{f}}}} 
\end{array}} \right] \hfill \\
   = \left[ {\begin{array}{*{20}{c}}
  {{{\mathbf{J}}_{{\text{ae}}}}^{\text{T}}}&{{{\mathbf{0}}_{3 \times 3n}}} \\ 
  {{{\mathbf{J}}_{{\text{af}}}}^{\text{T}}}&{{{\mathbf{I}}_{3n \times 3n}}} 
\end{array}} \right]\left[ {\begin{array}{*{20}{c}}
  {{{\mathbf{\tau }}_{\text{a}}}} \\ 
  {{{\mathbf{\tau }}_{\text{f}}}} 
\end{array}} \right]. \hfill \\ 
\end{gathered}
\end{equation}

\section{Mode shape selection}
In the previous section, the kinematics analysis and the dynamic modeling of the 3-\underline{R}RR PMFAL are complicated, but the mode shape functions have not been determined. The selection of mode shape function for a given flexible manipulator is not a clearly answered problem\cite{ref31}. The experimental validation is a common way to select mode shape functions, but it is accompanied by a huge cost. In contrast, getting data through commercial software is a cost-effective means.
\begin{table}
\begin{center}
\caption{Specification of the 3-\underline{R}RR PMFAL.}
\label{tab1}
\begin{tabular}{| c | c | c | c |}
\hline
Components & Parameters & Symbols & Values\\
\hline
Fixed base& Side length & R & 800 (mm)\\
\hline
Moving platform& Side length & r & 289 (mm)\\
\hline
Moving platform& Mass & $m_e$ & 0.83 (Kg)\\
\hline
Moving platform& Rotational inertia & $J_e$ & 5764 (${{\text{Kg}} \cdot {\text{m}}{{\text{m}}^2}}$)\\
\hline
Links& Length & $l_1/l_2$ & 600 (mm)\\
\hline
Links& Width & a & 30 (mm)\\
\hline
Links& Thickness & a & 5 (mm)\\
\hline
Links& Mass per unit length & ${\rho _i}$ & 0.4155 (${{{{\text{Kg}}} \mathord{\left/
 {\vphantom {{{\text{Kg}}} {\text{m}}}} \right.
 \kern-\nulldelimiterspace} {\text{m}}}}$)\\
 \hline
Links& Young’s modulus & E & $7.102 \times {10^{10}}{\text{ }}\left( {{\text{Pa}}} \right)$\\
\hline 
\end{tabular}
\end{center}
\end{table}

In this section, the multi-body software MSC ADAMS is used to simulate the dynamic behavior of the 3-\underline{R}RR PMFAL. At first, the geometrical and physical parameters of the 3-\underline{R}RR PMFAL are given in Table~\ref{tab1}. It should be noted that the actuation links and the intermediate links have the same parameters. Utilizing the View/Flex module of ADAMS, the actuation links are divided into finite elements, and the mesh modal of each actuation link has 9581 nodes and 4436 elements. In the simulation, the actuation links of the 3-\underline{R}RR PMFAL rotate 5 degrees counterclockwise in a short period of time (0.2 seconds), simulating a high-speed working condition.

Based on the above preparations, the algorithm of dynamic mode decomposition (DMD) is used to obtain the mode shape function of the flexible links under the high-speed working condition. DMD is inherently data-driven, and it starts with collecting a number of snapshots of the system states as they evolve in time\cite{ref32}. Firstly, we uniformly select nine sampling points on each flexible link to obtain the deformation data of the mechanism. A snapshot is the deformed state of the selected points at a specific moment and is reshaped into a nine-dimensional column vector. These snapshots are arranged into two data matrices, i.e., ${\mathbf{Y}}$ and ${\mathbf{Y'}}$, as follows
\begin{figure*}[!t]
\centering
\includegraphics[width=6in]{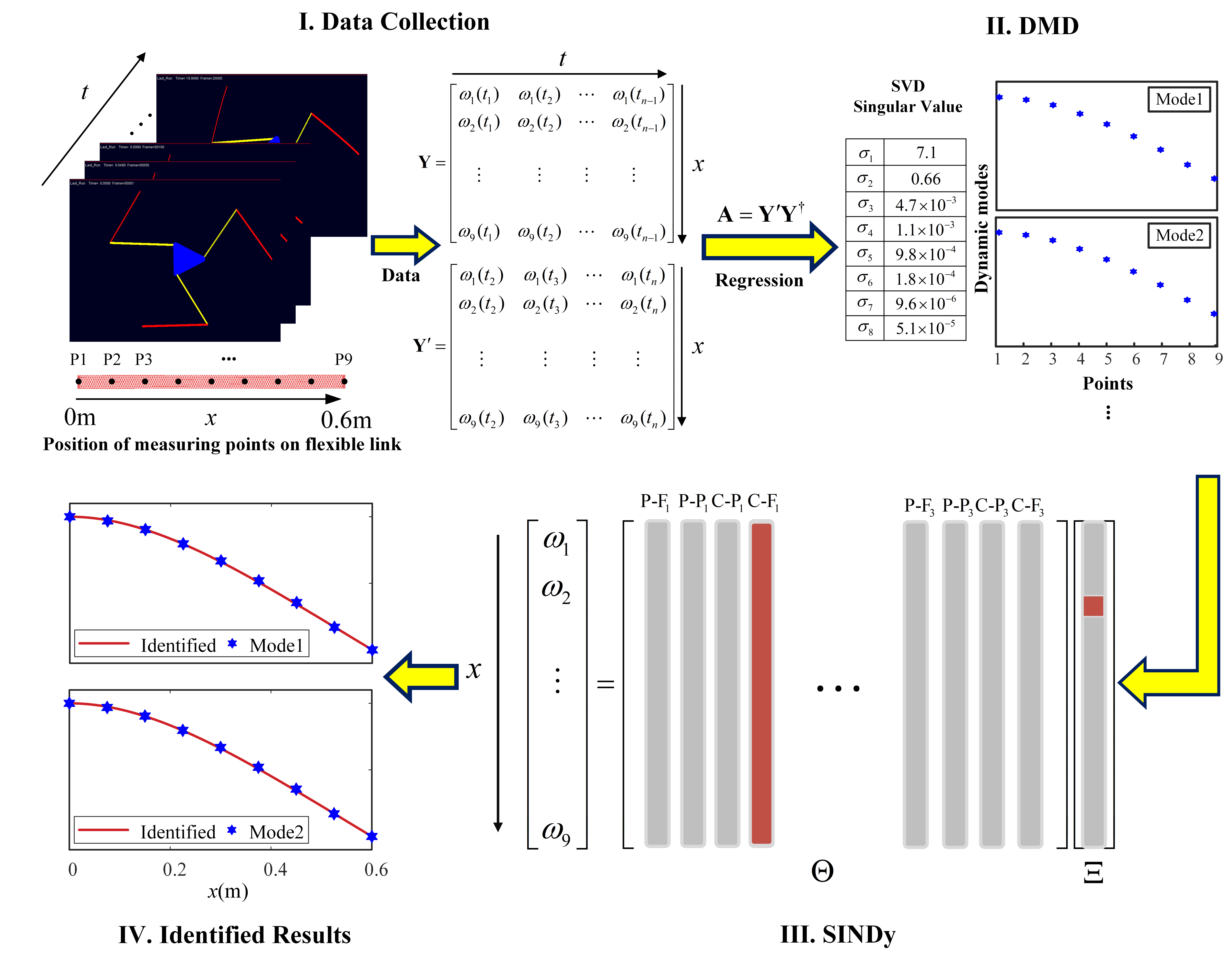}
\caption{Schematic of the selection of the mode shape function of the flexible links. The first step is to collect 20000 snapshots of the states of nine sampling points on each flexible link through a 20-second simulation in ADAMS. Next, the sampling snapshots are arranged into two data matrices, ${\mathbf{Y}}$ and ${\mathbf{Y'}}$, and the DMD algorithm is used to extract the mode shape function of the flexible links from the deformation data. Then, the extracted mode shape function is identified by SINDy algorithm. The results of sparse regression show that the fourth element of the coefficient vector ${\mathbf{\Xi }}$ is 0.454, and the other elements are close to zero. The effectiveness of the identification of SINDy is shown in IV.}
\label{fig_3}
\end{figure*}
\begin{equation}
\label{eq_34}
{\mathbf{Y}} = \left[ {\begin{array}{*{20}{c}}
  {\left| {} \right.}&{\left| {} \right.}&{}&{\left| {} \right.} \\ 
  {{\mathbf{\omega }}\left( {{t_1}} \right)}&{{\mathbf{\omega }}\left( {{t_2}} \right)}& \cdots &{{\mathbf{\omega }}\left( {{t_{n - 1}}} \right)} \\ 
  {\left| {} \right.}&{\left| {} \right.}&{}&{\left| {} \right.} 
\end{array}} \right],
\end{equation}
\begin{equation}
\label{eq_35}
{\mathbf{Y'}} = \left[ {\begin{array}{*{20}{c}}
  {\left| {} \right.}&{\left| {} \right.}&{}&{\left| {} \right.} \\ 
  {{\mathbf{\omega }}\left( {{t_2}} \right)}&{{\mathbf{\omega }}\left( {{t_3}} \right)}& \cdots &{{\mathbf{\omega }}\left( {{t_n}} \right)} \\ 
  {\left| {} \right.}&{\left| {} \right.}&{}&{\left| {} \right.} 
\end{array}} \right],
\end{equation}
where ${\mathbf{\omega }}\left( {{t_k}} \right) = {\left[ {\begin{array}{*{20}{c}}
  {{\omega _1}\left( {{t_k}} \right)}&{{\omega _2}\left( {{t_k}} \right)}& \cdots &{{\omega _9}\left( {{t_k}} \right)} 
\end{array}} \right]^{\text{T}}}$, and ${t_k} = k\Delta t$. The timestep $\Delta t$ is 0.001 seconds, which is sufficiently small to resolve the high frequencies in the dynamics. The DMD algorithm seeks the leading spectral decomposition of the best-fit linear operator ${\mathbf{A}}$ that relates the two snapshot matrices in time:
\begin{equation}
\label{eq_36}
{\mathbf{Y'}} \approx {\mathbf{AY}}.
\end{equation}
Mathematically, the best-fit operator ${\mathbf{A}}$ is given by
\begin{equation}
\label{eq_37}
{\mathbf{A}} = \mathop {{\text{argmin}}}\limits_{\mathbf{A}} {\left\| {{\mathbf{Y'}} - {\mathbf{AY}}} \right\|_F} = {\mathbf{Y'}}{{\mathbf{Y}}^\dag },
\end{equation}
where ${\left\| {{\text{ }} \cdot {\text{ }}} \right\|_F}$ is the Frobenius norm\cite{ref33}, and ${{\mathbf{Y}}^\dag }$ denotes the pseudo-inverse of ${{\mathbf{Y}}}$. By studying the operator ${{\mathbf{A}}}$, the mode shape function of the flexible links can be extracted and is denoted by ${{\mathbf{\omega }}_e}$. The detailed calculation process of the DMD algorithm can be referred to Ref.\cite{ref34}.

Next, the algorithm of the sparse identification of nonlinear dynamics (SINDy) is applied to identify the mode shape function, namely, retrieve the mode shape as a function of the flexible links out of the discrete data from DMD.  As for SINDy, the most important part is the construction of the library ${\mathbf{\Theta }}$. Herein, we include in the library ${\mathbf{\Theta }}$ the modal shape functions of first three orders under the following boundary conditions: Free-Free, Pinned-Pinned, Pinned-Free, Clamped-Free boundary condition, etc. Then ${{\mathbf{\omega }}_e}$ can be represented in terms of ${\mathbf{\Theta }}$ as
\begin{equation}
\label{eq_38}
{{\mathbf{\omega }}_e} = {\mathbf{\Theta \Xi }},
\end{equation}
where ${\mathbf{\Xi }}$ is a vector of coefficients determining the active terms of the library ${\mathbf{\Theta }}$. To improve the numerical robustness of the identification for noisy data, the sparse regression method, i.e., the LASSO algorithm\cite{ref35}, is used in realizing SINDy. Then ${\mathbf{\Xi }}$ is defined as
\begin{equation}
\label{eq_39}
{\mathbf{\Xi }} = \mathop {{\text{argmin}}}\limits_{\mathbf{\Xi }} {\left\| {{{\mathbf{\omega }}_e} - {\mathbf{\Theta \Xi }}} \right\|_2} + \lambda {\left\| {\mathbf{\Xi }} \right\|_1},
\end{equation}
where $\lambda$ is a coefficient used to adjust the sparsity and accuracy of the LASSO algorithm. The calculation process of SINDy can be referred to Ref.\cite{ref36}.

\begin{table}
\begin{center}
\caption{Range of the inputs of the training data for the neural-network-based state observer.}
\label{tab2}
\begin{tabular}{| c | c |}
\hline
Description & Range\\
\hline
Range of $x$ (mm)& $\left[ { - 150,{\text{ }}150} \right]$\\
\hline
Range of $y$ (mm)& $\left[ { - 150,{\text{ }}150} \right]$\\
\hline
Range of $\theta$ (rad)& $\left[ { - 0.2,{\text{ }}0.2} \right]$\\
\hline
Range of ${\left. {{\omega _1}} \right|_{x = l}}$ (mm)& $\left[ { - 60,{\text{ }}60} \right]$\\
\hline
Range of ${\left. {{\omega _2}} \right|_{x = l}}$ (mm)& $\left[ { - 60,{\text{ }}60} \right]$\\
\hline
Range of ${\left. {{\omega _3}} \right|_{x = l}}$ (mm)& $\left[ { - 60,{\text{ }}60} \right]$\\
\hline 
\end{tabular}
\end{center}
\end{table}

The process of the selection of mode shape function is shown in Fig. \ref{fig_3}. The results show that the identified modal shape function is in good agreement with ${{\mathbf{\omega }}_e}$, and the active term of ${\mathbf{\Xi }}$ corresponds to ${\text{C - }}{{\text{F}}_1}$ (III in Fig. \ref{fig_3}), which indicates that the modal shape function of the flexible links of 3-\underline{R}RR PMFAL is consistent with the first-order mode shape function of the Clamped-Free boundary condition. Therefore, the mode shape function ${\phi _{ij}}\left( x \right)$ in Eq. (4) is given by
\begin{equation}
\label{eq_40}
{\phi _{ij}}\left( x \right) = \cos \left( {{\beta _j}x} \right) - \cosh \left( {{\beta _j}x} \right) + {r_j}\left[ {\sin \left( {{\beta _j}x} \right) - \sinh \left( {{\beta _j}x} \right)} \right],
\end{equation}
where ${r_j} = {{\left[ {\sin \left( {{\beta _j}x} \right) - \sinh \left( {{\beta _j}x} \right)} \right]} \mathord{\left/
 {\vphantom {{\left[ {\sin \left( {{\beta _j}x} \right) - \sinh \left( {{\beta _j}x} \right)} \right]} {\left[ {\cos \left( {{\beta _j}x} \right) + \cosh \left( {{\beta _j}x} \right)} \right]}}} \right.
 \kern-\nulldelimiterspace} {\left[ {\cos \left( {{\beta _j}x} \right) + \cosh \left( {{\beta _j}x} \right)} \right]}}.$

\begin{figure}[!t]
\centering
\includegraphics[width=3in]{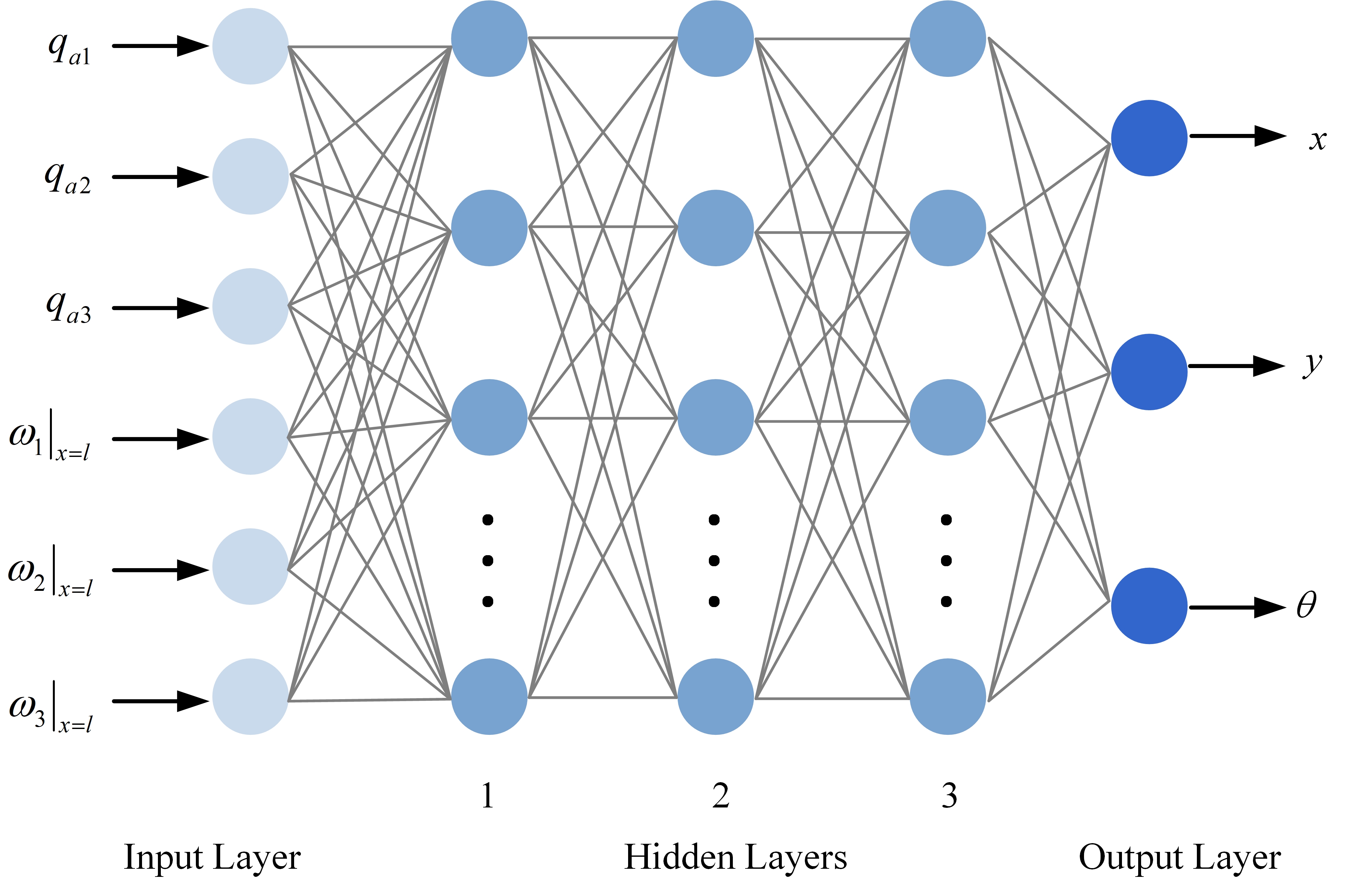}
\caption{The structure of the neural network. To balance the fitting accuracy and calculation efficiency of the neural network, the number of hidden layers is chosen as 3, and the numbers of the neurons in each hidden layer are 30, 30, and 30, respectively.}
\label{fig_4}
\end{figure}
 
\section{Model-based control with state feedback}
In the above sections, the dynamic model of the 3-\underline{R}RR PMFAL is established and the boundary condition during the motion of the mechanism is determined, providing a promising basis for designing the model-based control law. Since the real-time state-feedback can effectively cope with the perturbations caused by the deformation of flexible links, it is necessary to construct a state observer of the mechanism, and then the model-based controller can be achieved by combining the real-time state observer.

\subsection{Neural-network-based state observer}
In the dynamic model, the generalized coordinates of the 3-\underline{R}RR PMFAL are chosen as ${\mathbf{q}} = {\left[ {\begin{array}{*{20}{c}}
  {{{\mathbf{q}}_{\text{e}}}}&{{{\mathbf{q}}_{\text{f}}}} 
\end{array}} \right]^{\text{T}}}$ where the mode coordinates ${{\mathbf{q}}_{\text{f}}}$ can be collected in real-time by experimental equipment such as dynamic strain acquisition system. However, due to the difficulties in analytically solving the forward kinematics of parallel manipulators, the coordinate of the end-effector, i.e., ${\mathbf{T}}$, is hardly available from the sensor data of motors and flexible links. Given the fact that the neural network has the extraordinary capability of approximation\cite{ref37}, we construct a neural-network-based state observer for the end-effector to predict in real-time the state of the end-effector.

To obtain the training data of the neural network, 100000 data points are randomly selected within a prescribed region whose range is shown in Table~\ref{tab2}. The angles of the drive motors $\left( {{q_{a1}},{\text{ }}{q_{a2}},{\text{ }}{q_{a3}}} \right)$ corresponding to the 100000 data points are calculated through the inverse kinematics of the mechanism. The input of the neural network is formulated as the vector consisting of the angles of the drive motors and the deformation at the tip of the flexible links, i.e., ${\left( {{{\left. {{q_{a1}},{\text{ }}{q_{a2}},{\text{ }}{q_{a3}},{\text{ }}{\omega _1}} \right|}_{x = {l_1}}},{\text{ }}{{\left. {{\omega _2}} \right|}_{x = {l_1}}},{\text{ }}{{\left. {{\omega _3}} \right|}_{x = {l_1}}}} \right)^{\text{T}}}$, and the output is the vector representing the position and orientation of the end-effector, i.e., $\left( {x,{\text{ }}y,\theta } \right)$. The structure of the neural network is shown in Fig. \ref{fig_4}, where the number of the hidden layers and the number of neurons in each hidden layer are a consequence of balancing the fitting accuracy and the calculation efficiency. After twenty hours of training, the mean square error (MSE) of the trained neural network on the test set has decreased to $1.33 \times {10^{ - 10}}$. In order to show the training performance of the neural network, another data set with 10000 points, viewed as a test set, is generated within the same region using the aforementioned method. After testing on the test set, the maximum errors and the MSEs of the predictions of the state observer are calculated and shown in Table 3, indicating that the prediction of the constructed state observer is effective. Meanwhile, the time taken for the neural-network-based state observer to make 10000 predictions is around 0.275 seconds on a computer with Intel (R) Core (TM) i7-6700K CPU @ 4.00 GHz, and 16 GB RAM, indicating the excellent efficiency of the online deployed state observer.
\begin{table}
\begin{center}
\caption{The performances of the trained neural network on the test set.}
\label{tab3}
\begin{tabular}{| c | c | c |}
\hline
 & Maximum error & MSE\\
\hline
$x$ (mm) & $2.15 \times {10^{ - 5}}$ & $3.68 \times {10^{ - 12}}$\\
\hline
$y$ (mm) & $9.38 \times {10^{ - 6}}$ & $3.00 \times {10^{ - 12}}$\\
\hline
$\theta$ (rad) & $2.60 \times {10^{ - 5}}$ & $1.33 \times {10^{ - 11}}$\\
\hline 
\end{tabular}
\end{center}
\end{table}

\subsection{Model-based control with neural-network-based state observer}
Based on the constructed state observer, the state of the end-effector of the 3-\underline{R}RR PMFAL can be obtained in real-time by the states of drive motors and flexible links. On this foundation, the model-based control with feedback compensation can be realized for trajectory tracking and vibration suppression of the mechanism.

\begin{figure*}[!t]
\centering
\includegraphics[width=5.5in]{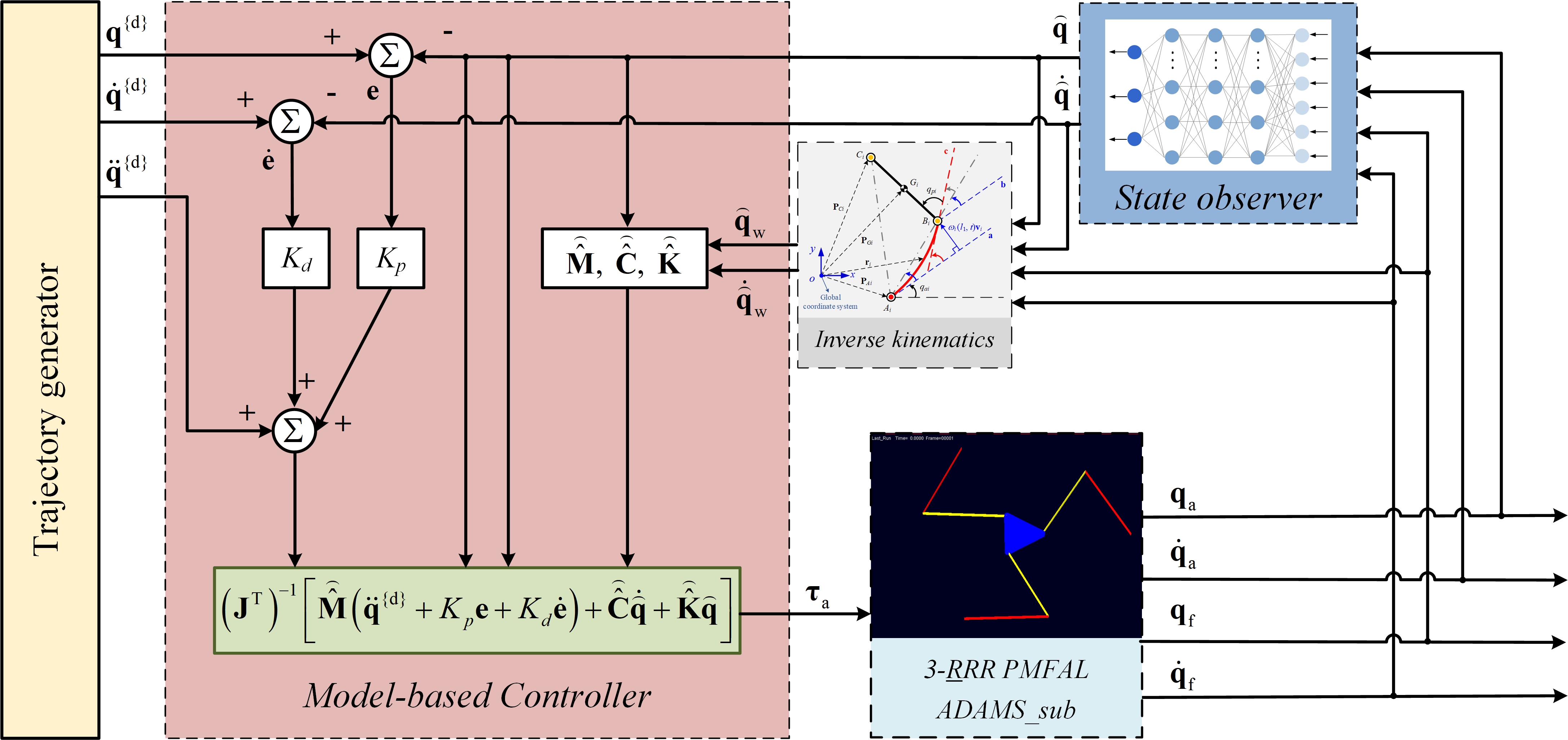}
\caption{Block diagram of model-based control with neural-network-based state observer.}
\label{fig_5}
\end{figure*}

Denote the coordinates of the end-effector obtained from the state observer by ${{\mathbf{\overset{\lower0.5em\hbox{$\smash{\scriptscriptstyle\frown}$}}{q} }}_{\text{e}}}$, and ${{\mathbf{\overset{\lower0.5em\hbox{$\smash{\scriptscriptstyle\frown}$}}{q} }}_{\text{e}}} = {\left[ {\begin{array}{*{20}{c}}
  {\overset{\lower0.5em\hbox{$\smash{\scriptscriptstyle\frown}$}}{x} }&{\overset{\lower0.5em\hbox{$\smash{\scriptscriptstyle\frown}$}}{y} }&{\overset{\lower0.5em\hbox{$\smash{\scriptscriptstyle\frown}$}}{\theta } } 
\end{array}} \right]^{\text{T}}}$. Combining the collected sensor data of flexible links, the estimates of the generalized coordinates are denoted by ${\mathbf{\overset{\lower0.5em\hbox{$\smash{\scriptscriptstyle\frown}$}}{q} }}$, and ${\mathbf{\overset{\lower0.5em\hbox{$\smash{\scriptscriptstyle\frown}$}}{q} }} = {\left[ {\begin{array}{*{20}{c}}
  {{\mathbf{\overset{\lower0.5em\hbox{$\smash{\scriptscriptstyle\frown}$}}{q} }}_{\text{e}}^{\text{T}}}&{{\mathbf{q}}_{\text{f}}^{\text{T}}} 
\end{array}} \right]^{\text{T}}}$. The tracking errors of the generalized coordinates of the 3-\underline{R}RR PMFAL are defined as
\begin{equation}
\label{eq_41}
{\mathbf{e}} = {{\mathbf{q}}^{\{ {\text{d}}\} }} - {\mathbf{\overset{\lower0.5em\hbox{$\smash{\scriptscriptstyle\frown}$}}{q} }},
\end{equation}
where ${\{  \cdot \} ^{\{ {\text{d}}\} }}$ denotes the desired signal of $\{  \cdot \} $. The deformation of the flexible links is undesired during the operation of the parallel mechanism, and thus the desired signal can be expressed as
\begin{equation}
\label{eq_42}
{{\mathbf{q}}^{\{ {\text{d}}\} }} = {\left[ {\begin{array}{*{20}{c}}
  {{x^{\{ {\text{d}}\} }}}&{{y^{\{ {\text{d}}\} }}}&{{\theta ^{\{ {\text{d}}\} }}}&{{{\mathbf{0}}_{1 \times 3n}}} 
\end{array}} \right]^{\text{T}}}.
\end{equation}
Taking advantage of the non-linear compensation, a classical model-based method named computed torque control is employed for the control law design. With that, the model-based control law with feedback compensation can be designed as
\begin{equation}
\label{eq_43}
\begin{aligned} & \boldsymbol{\tau}=\left[\begin{array}{ll}\boldsymbol{\tau}_{\mathrm{a}}{ }^{\mathrm{T}} & \boldsymbol{\tau}_{\mathrm{f}}{ }^{\mathrm{T}}\end{array}\right]^{\mathrm{T}} \\ & =\left(\mathbf{J}^{\mathrm{T}}\right)^{-1}\left[{{\mathbf{\overset{\lower0.5em\hbox{$\smash{\scriptscriptstyle\frown}$}}{\hat M} }}}\left(\ddot{\mathbf{q}}^{\{\mathrm{d}\}}+K_p \mathbf{e}+K_d \dot{\mathbf{e}}\right)+{{\mathbf{\overset{\lower0.5em\hbox{$\smash{\scriptscriptstyle\frown}$}}{\hat C} }}} \dot{\overline{\mathbf{q}}}+{{\mathbf{\overset{\lower0.5em\hbox{$\smash{\scriptscriptstyle\frown}$}}{\hat K} }}} \hat{\mathbf{q}}\right],\end{aligned}
\end{equation}
where $K_p$ and $K_d$ respectively represent the position gain and velocity gain. In Eq.(\ref{eq_43}), ${{\mathbf{\overset{\lower0.5em\hbox{$\smash{\scriptscriptstyle\frown}$}}{\hat M} }}}$, ${{\mathbf{\overset{\lower0.5em\hbox{$\smash{\scriptscriptstyle\frown}$}}{\hat C} }}}$ and ${{\mathbf{\overset{\lower0.5em\hbox{$\smash{\scriptscriptstyle\frown}$}}{\hat K} }}}$ represent the estimates of ${\mathbf{\hat M}}$, ${\mathbf{\hat C}}$ and ${\mathbf{\hat K}}$, respectively, and are generated by replacing ${\mathbf{q}}$ and ${\mathbf{\dot q}}$ by their estimates in ${\mathbf{\hat M}}$, ${\mathbf{\hat C}}$ and ${\mathbf{\hat K}}$.  The stability analysis of the model-based controller is presented. At first, a theorem about the state observer will be introduced as follows.

\textbf{Theorem 1}. Multi-layer feedforward neural networks with an appropriately smooth hidden layer activation function are able to approximate any function and its derivatives with arbitrary accuracy\cite{ref38}.

The theorem means that by selecting a neural network with reasonable structure and sufficient training time, the estimates obtained by the neural-network-based state observer can predict the real state of the system to any desired accuracy. Ignoring the error of data acquisition, the prediction of the state observer can be expressed as
\begin{equation}
\label{eq_44}
\lim {\mathbf{\overset{\lower0.5em\hbox{$\smash{\scriptscriptstyle\frown}$}}{T} }} \to {\mathbf{T}}.
\end{equation}
Combing Eqs. (\ref{eq_32}), (\ref{eq_43}) and (\ref{eq_44}), the dynamic equation with the proposed control law can be written as
\begin{equation}
\label{eq_45}
\begin{gathered}
  {\mathbf{\hat M\ddot q}} + {\mathbf{\hat C\dot q}} + {\mathbf{\hat Kq}} =  \hfill \\
  {{\mathbf{J}}^{\text{T}}}{\left( {{{\mathbf{J}}^{\text{T}}}} \right)^{ - 1}}\left[ {{\mathbf{\hat M}}\left( {{{{\mathbf{\ddot q}}}^{\{ {\text{d}}\} }} + {K_p}{\mathbf{e}} + {K_d}{\mathbf{\dot e}}} \right) + {\mathbf{\hat C\dot q}} + {\mathbf{\hat Kq}}} \right], \hfill \\ 
\end{gathered}
\end{equation}
Eq. (\ref{eq_45}) can be reduced to
\begin{equation}
\label{eq_46}
{\mathbf{\hat M}}\left( {{\mathbf{\ddot e}} - {K_p}{\mathbf{e}} - {K_d}{\mathbf{\dot e}}} \right) = {\mathbf{0}},
\end{equation}
which is equivalent to
\begin{equation}
\label{eq_47}
{\mathbf{\ddot e}} = {K_p}{\mathbf{e}} + {K_d}{\mathbf{\dot e}}.
\end{equation}
With the above preparation, the Lyapunov function is defined as
\begin{equation}
\label{eq_48}
V = \frac{1}{2}{{\mathbf{\dot e}}^{\text{T}}}{\mathbf{\dot e}} + \frac{1}{2}{K_p}{{\mathbf{e}}^{\text{T}}}{\mathbf{e}}.
\end{equation}
Differentiating both sides of Eq. (\ref{eq_48}) with respect to time yields
\begin{equation}
\label{eq_49}
\begin{gathered}
  \dot V = {{{\mathbf{\dot e}}}^{\text{T}}}{\mathbf{\ddot e}} + {K_p}{{{\mathbf{\dot e}}}^{\text{T}}}{\mathbf{e}} \hfill \\
   =  - {{{\mathbf{\dot e}}}^{\text{T}}}\left( {{K_p}{\mathbf{e}} + {K_d}{\mathbf{\dot e}}} \right) + {K_p}{{{\mathbf{\dot e}}}^{\text{T}}}{\mathbf{e}} \hfill \\
   =  - {K_d}{{{\mathbf{\dot e}}}^{\text{T}}}{\mathbf{\dot e}}. \hfill \\ 
\end{gathered} 
\end{equation}
It can be seen from Eq. (\ref{eq_49}) that $\dot V \leq 0$ when ${K_d} > 0$. In this case, the lower bound of the constructed Lyapunov function is zero, which ensures the stability of the model-based control law.

 Since the distributed smart materials such as the piezoelectric transducer (PZT) are difficult to implement, a manipulator with flexible links is generally considered as an under-actuated system\cite{ref39}. In this case, the controllable variables of the 3-RRR PMFAL consist of only the three active joints. Based on Eqs. (\ref{eq_33}) and (\ref{eq_43}), ${{\mathbf{\tau }}_{\text{a}}}$ in partitioned form can be written as
\begin{equation}
\label{eq_50}
{{\mathbf{\tau }}_{\text{a}}} = {\left( {{{\mathbf{J}}_{{\text{ax}}}}^{\text{T}}} \right)^{ - 1}}\left( \begin{gathered}
  {{{\mathbf{\overset{\lower0.5em\hbox{$\smash{\scriptscriptstyle\frown}$}}{\hat M} }}}_{{\text{rr}}}}{\mathbf{\ddot {\overset{\lower0.5em\hbox{$\smash{\scriptscriptstyle\frown}$}}{q} }}}_{\text{e}}^{\{ {\text{d}}\} } + {K_p}{{{\mathbf{\overset{\lower0.5em\hbox{$\smash{\scriptscriptstyle\frown}$}}{\hat M} }}}_{{\text{rr}}}}{{\mathbf{e}}_{\text{e}}} + {K_p}{{{\mathbf{\overset{\lower0.5em\hbox{$\smash{\scriptscriptstyle\frown}$}}{\hat M} }}}_{{\text{rf}}}}{{\mathbf{e}}_{\text{f}}} +  \hfill \\
  {K_d}{{{\mathbf{\overset{\lower0.5em\hbox{$\smash{\scriptscriptstyle\frown}$}}{\hat M} }}}_{{\text{rr}}}}{{{\mathbf{\dot e}}}_{\text{e}}} + {K_d}{{{\mathbf{\overset{\lower0.5em\hbox{$\smash{\scriptscriptstyle\frown}$}}{\hat M} }}}_{{\text{rf}}}}{{{\mathbf{\dot e}}}_{\text{f}}} + {{{\mathbf{\overset{\lower0.5em\hbox{$\smash{\scriptscriptstyle\frown}$}}{\hat C} }}}_{{\text{rr}}}}{{{\mathbf{\dot q}}}_{\text{e}}} + {{{\mathbf{\overset{\lower0.5em\hbox{$\smash{\scriptscriptstyle\frown}$}}{\hat C} }}}_{{\text{rf}}}}{{{\mathbf{\dot q}}}_{\text{f}}} \hfill \\ 
\end{gathered}  \right),
\end{equation}
where ${{\mathbf{e}}_{\text{e}}} = {\mathbf{q}}_{\text{e}}^{\{ {\text{d}}\} } - {{\mathbf{\overset{\lower0.5em\hbox{$\smash{\scriptscriptstyle\frown}$}}{q} }}_{\text{e}}}$ and ${{\mathbf{e}}_{\text{f}}} = {\mathbf{q}}_{\text{f}}^{\{ {\text{d}}\} } - {{\mathbf{q}}_{\text{f}}}$. The block diagram of the model-based control is shown in Fig. 5, where the control law is deployed via SIMULINK and the forward dynamics calculation of the 3-\underline{R}RR PMFAL is performed in ADAMS. The SIMULINK/ADAMS co-simulation enables not only illustrating the vibration suppression effect of the proposed controller, but also verifying the reliability of the developed dynamics model by model compensation.

\begin{figure*}[!t]
\centering
\includegraphics[width=5in]{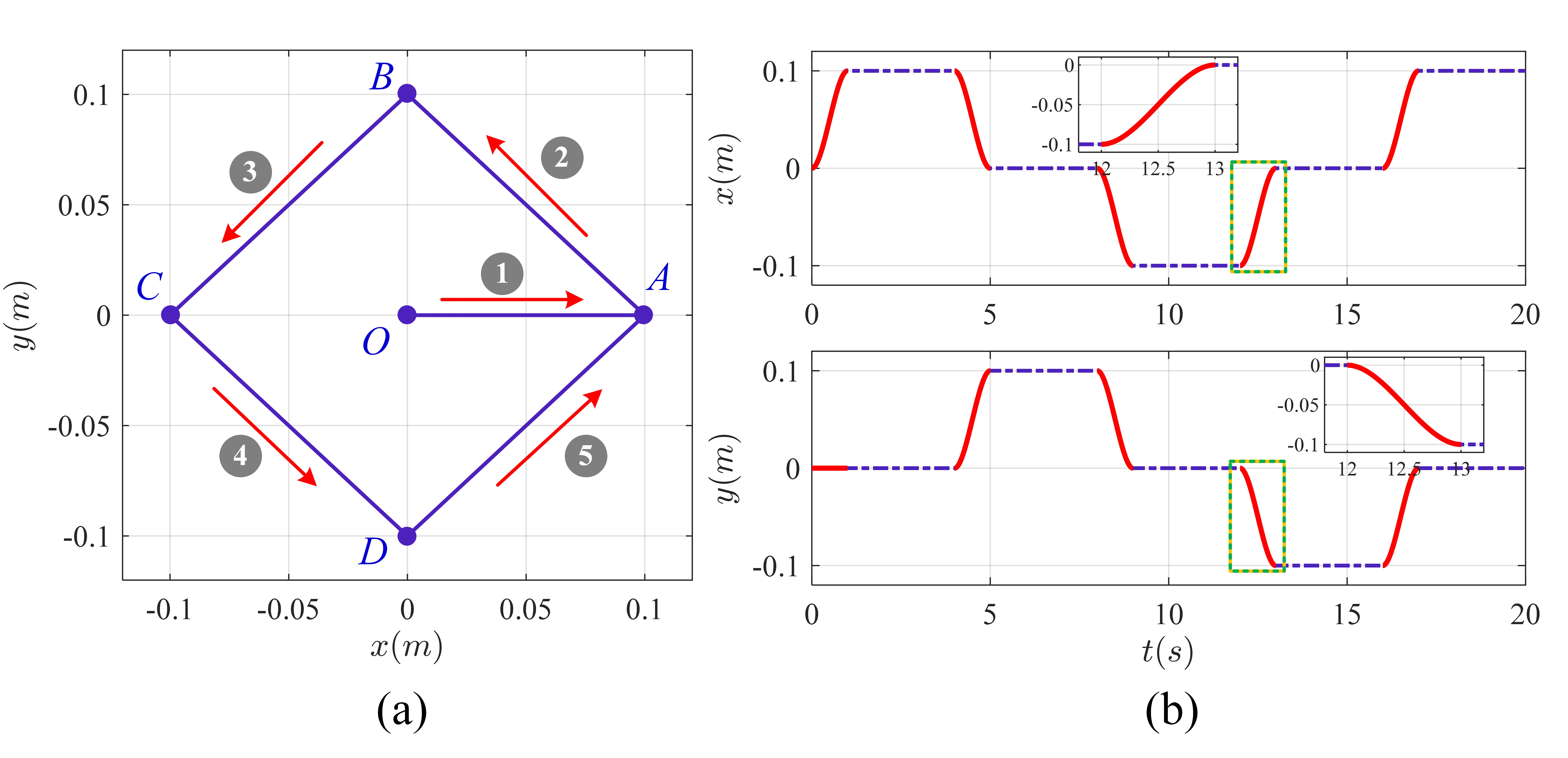}
\caption{Diagram of the trajectory of rapid positioning. (a) The trajectory between five points. (b) The trajectory time history of the end-effector in the x and y directions and two red curves are zoomed in the figure.}
\label{fig_6}
\end{figure*}

\section{Case study and results discussion}
In order to test the effect of the proposed controller, the working condition for rapid positioning of the 3-\underline{R}RR PMFAL is introduced in this part. As shown in Fig. \ref{fig_6}(a), a trajectory traversing the five points, i.e., O, A, B, C, and D, is illustrated, while the angle of rotation of the end-effector remains zero. The time history of the trajectory of the end-effector in the x and y directions is shown in Fig. \ref{fig_6}(b). In Fig. \ref{fig_6}(b), the red curves generated by cubic interpolation represent moving phases with length of 1 second, and the blue lines represent stabilization phases with length of 3 seconds.

\textbf{Remark 1}. The modes are truncated at the third order during model compensation in the co-simulation. Based on the analysis of the simulation results in ADAMS, the modes of first three orders are dominant in the vibration of the 3-\underline{R}RR PMFAL. Therefore, it is accurate enough to select the first three modes to represent the vibration response.

In addition, to validate the performance of the proposed control strategy, the joint-based PD control is employed for the above trajectory tracking as a comparison. For the proposed controller, the proportional gain is chosen as ${K_p} = 200$, and the differential gain is chosen as ${K_d} = 1$. For the joint-based PD control, the proportional gain is chosen as ${K_p} = 200$, and the differential gain is chosen as ${K_d} = 0.2$ (the greater differential gain of the joint-based PD control will lead to simulation divergence). Meanwhile, the geometrical and physical parameters of the 3-\underline{R}RR PMFAL are consistent with Table \ref{tab1}.

\begin{figure}[!t]
\centering
\includegraphics[width=3.4in]{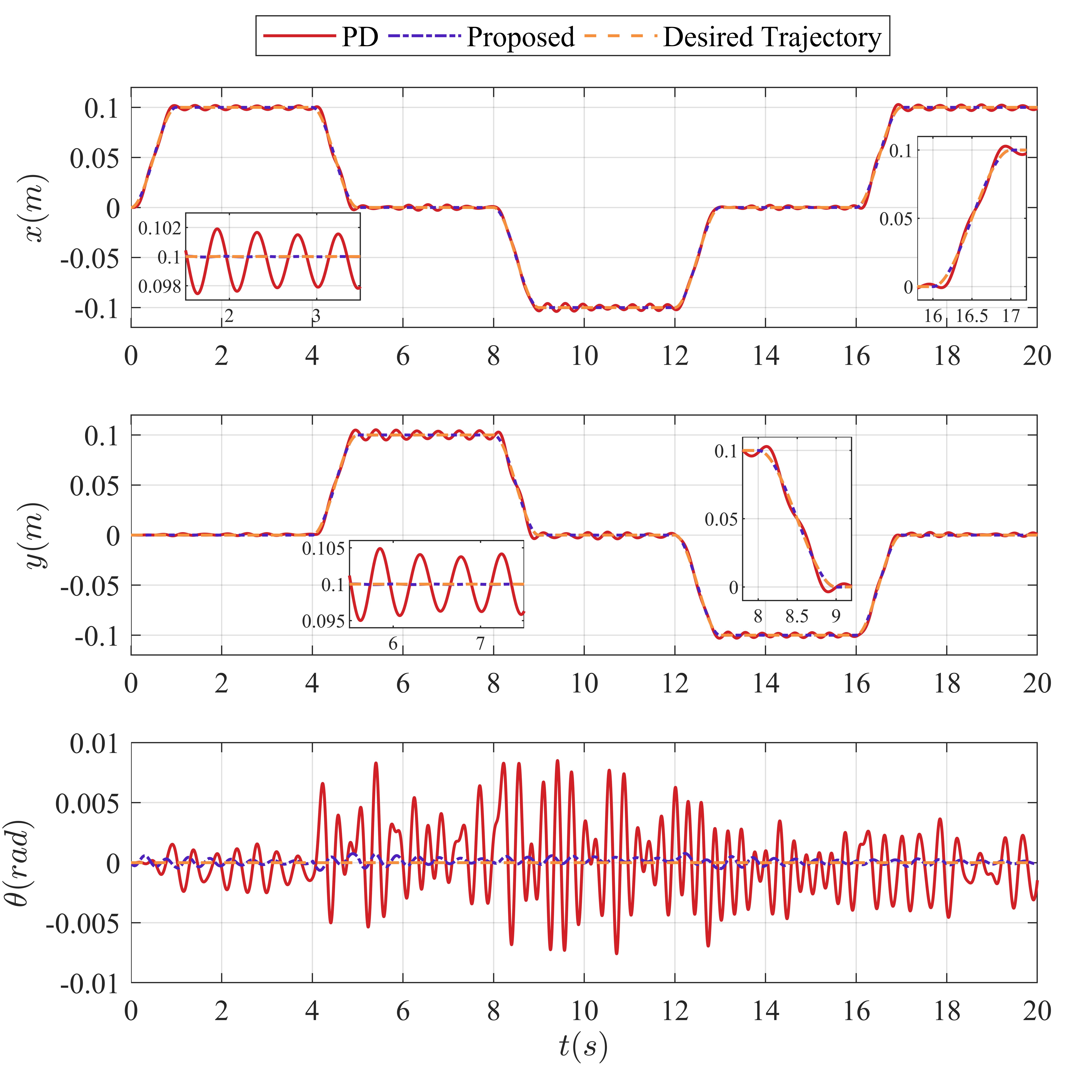}
\caption{The comparison of the trajectory tracking between the joint-based PD control and the proposed control. The states of the three degrees of freedom of the end-effector under two control strategies are shown in sequence, and stabilization phases and moving phases in the x-direction and the y-direction are amplified respectively.}
\label{fig_7}
\end{figure}

\subsection{Performance comparison between the joint-based controller and the proposed controller}
\begin{figure}[!t]
\centering
\includegraphics[width=3.4in]{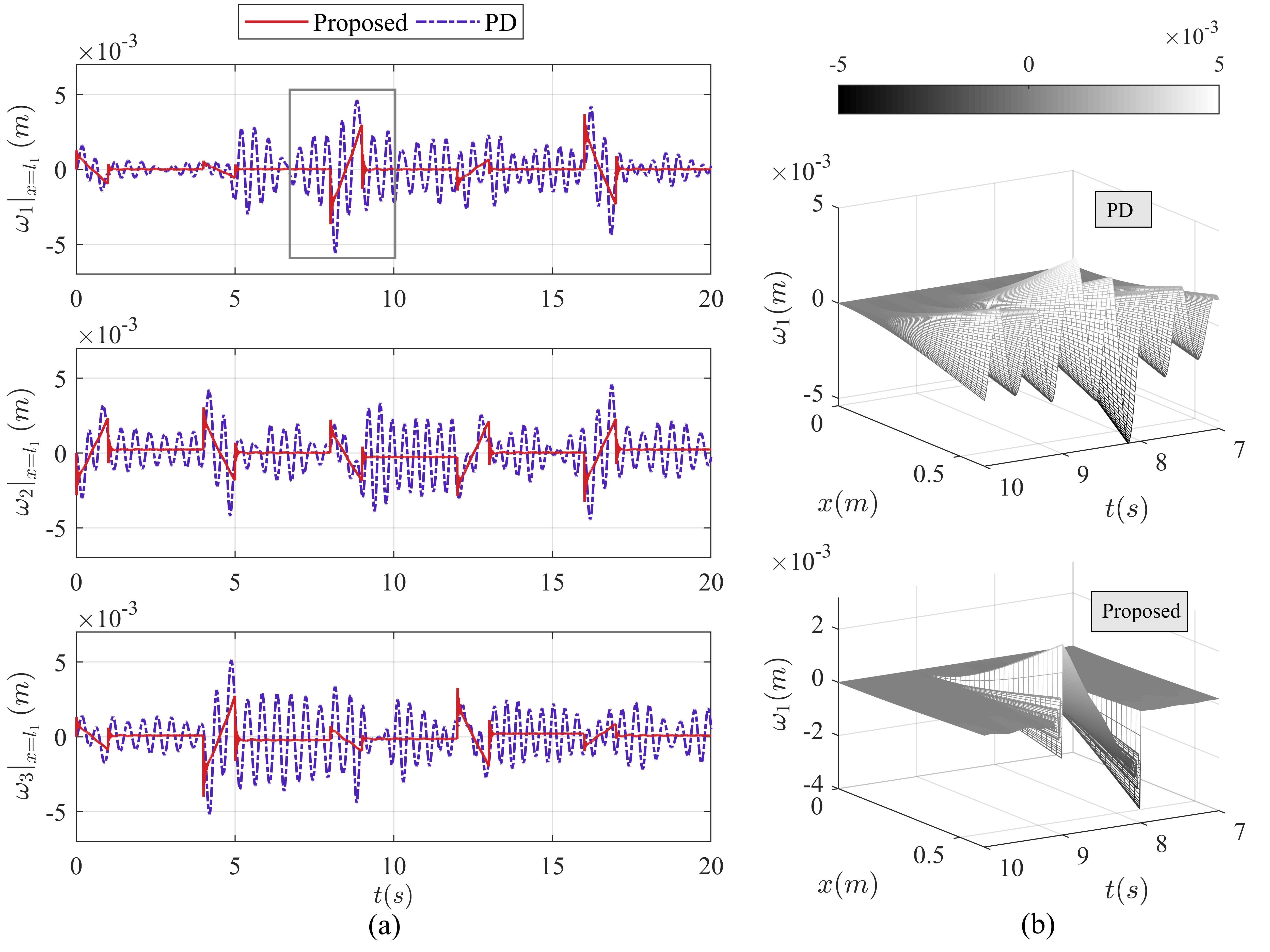}
\caption{The comparison of the deformation of three flexible links between the proposed control and the joint-based PD control in (a) the end-point deformations of three flexible links and (b) the flexural deformation of the entire flexible link in the time span corresponding to the grey box in the left column.}
\label{fig_8}
\end{figure}

\begin{figure}[!t]
\centering
\includegraphics[width=3.4in]{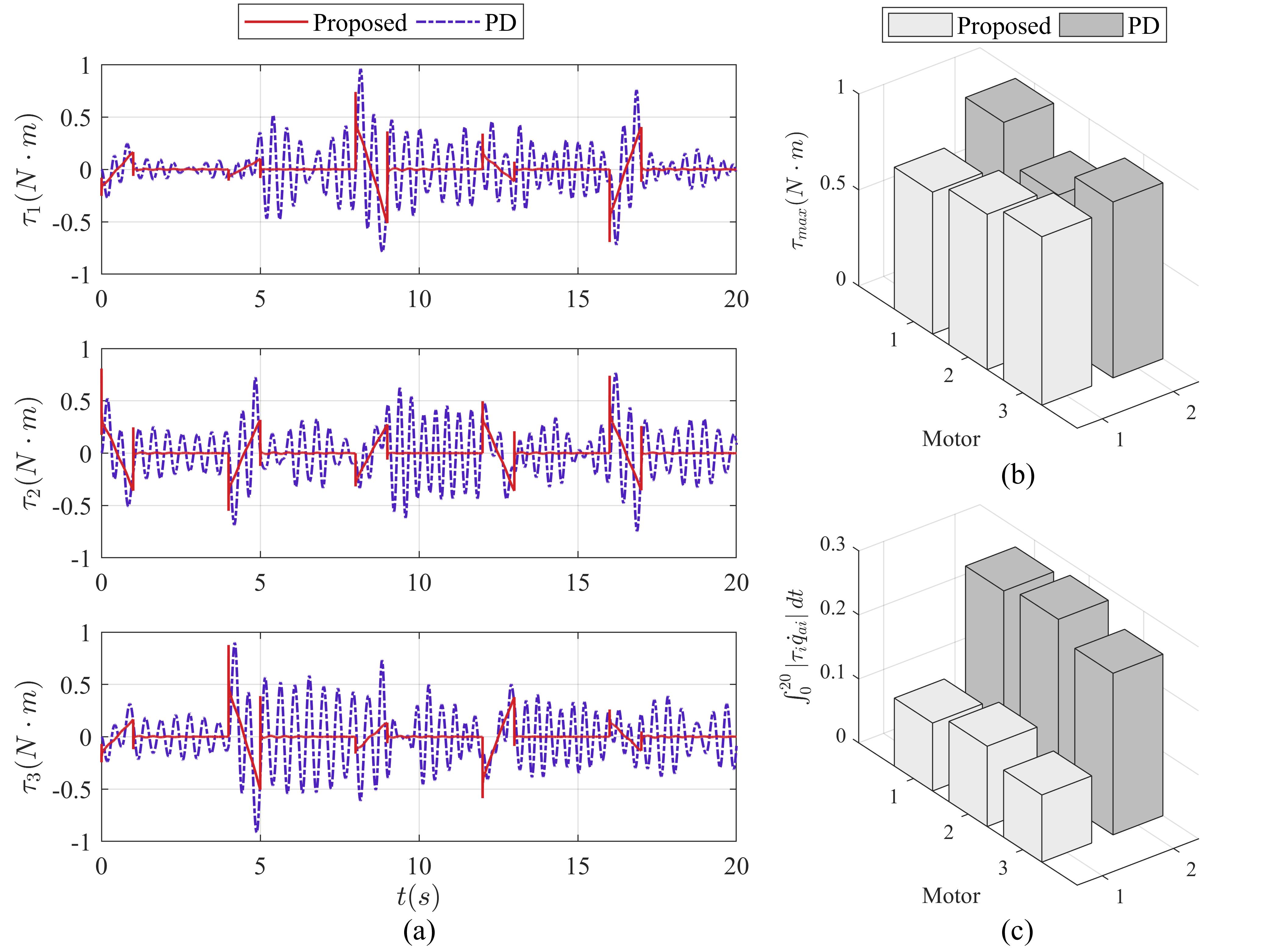}
\caption{The comparison of the torque inputs of the drive motors between the proposed control and the joint-based PD control in (a) time history of the torque inputs, (b) peak values of the torque inputs, and (c) energy consumption of the drive motors.}
\label{fig_9}
\end{figure}

In this section, the feedback frequency of the state observer and control frequency of the three motors are both chosen as 1000Hz. Then, the trajectory tracking, the deformation of three flexible links, and the torque inputs of two different control strategies are shown in Figs. \ref{fig_7}, \ref{fig_8}, and \ref{fig_9}, respectively.


As shown in Fig.\ref{fig_7}, regardless of the phases, the proposed control can realize a better tracking for desired trajectory in comparison with the joint-based PD control. As can be seen from the enlarged images, the tracking error of the proposed controller has been dropped to the order of $1 \times {10^{ - 4}}{\text{ }}m$ during 1.5 s to 3.5 s in the x-direction and 5.5 s to 7.5 s in the y-direction, while the tracking error of the joint-based PD control is still at the order of $1 \times {10^{ - 3}}{\text{ }}m$.

Since the Clamped-Free boundary condition of flexible links is selected, the end-point deformations of three flexible links have significant impact on the end-effector and consequently we use ${\left. {{\omega _1}} \right|_{x = {l_1}}}$, ${\left. {{\omega _2}} \right|_{x = {l_1}}}$, and ${\left. {{\omega _3}} \right|_{x = {l_1}}}$ to reflect the deformation of the overall flexible links. As shown in Fig. \ref{fig_8}(a), the end-point deformations under the proposed controller have been dropped rapidly to around zero during stabilization phases, but that under the joint-based PD control still have considerable oscillations. 

The proposed controller has good performance not only in the above two aspects but also in the torque inputs of the drive motors. As shown in Fig. \ref{fig_9}(a), the torque inputs under both control strategies show significant jitter in the moving phases, but the torque inputs of the proposed controller drop faster than that of the joint-based PD control in the stabilization phases. The peak values of the torque inputs have identical magnitude for both control methods, as shown in Fig. \ref{fig_9}(b), while the proposed controller manifests its advantage in energy consumption of the drive motors as shown in Fig. \ref{fig_9}(c).

\begin{figure}[!t]
\centering
\includegraphics[width=3.2in]{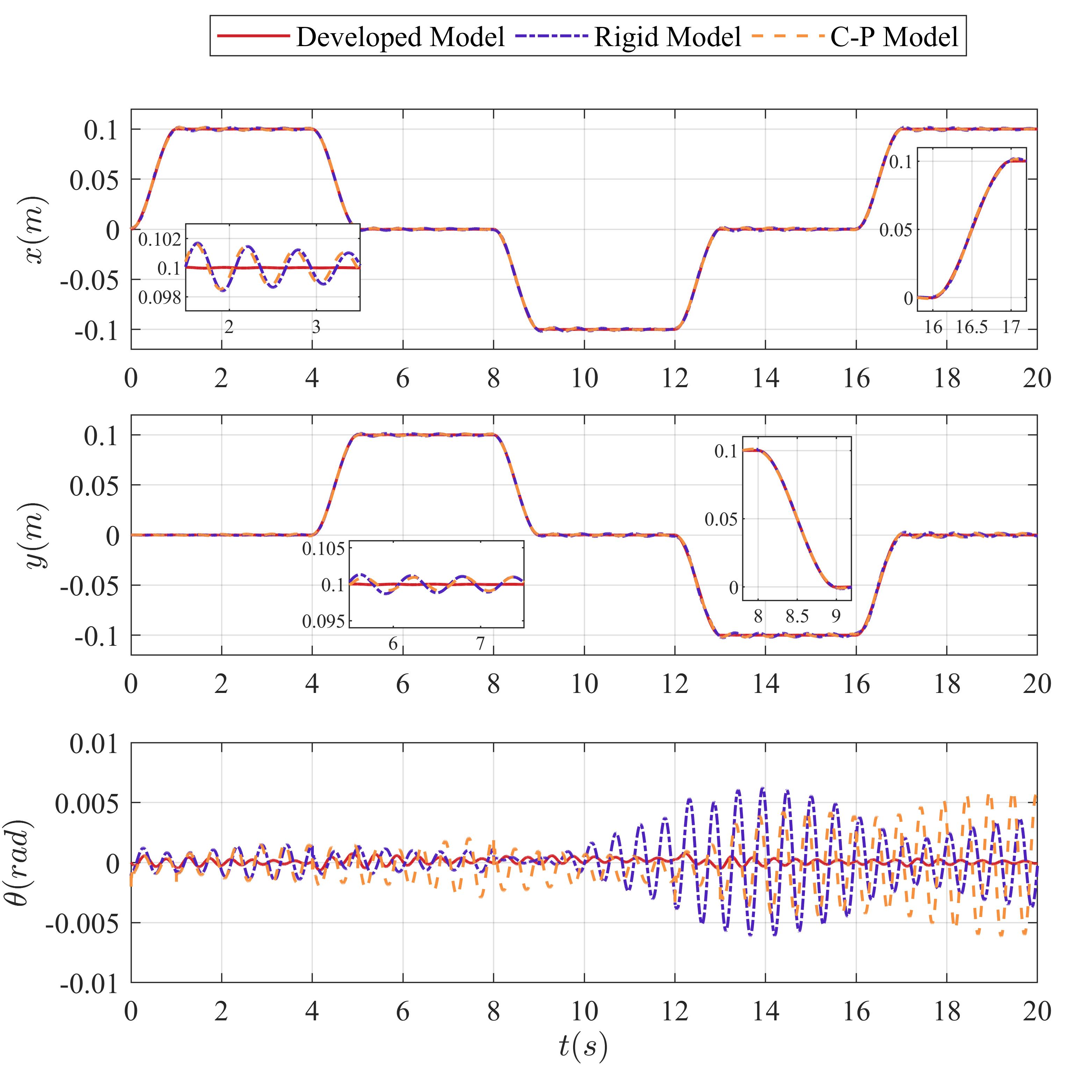}
\caption{Trajectory tracking results for the proposed controller with different dynamic models, revealing lower tracking error under the proposed controller with developed model, and thus demonstrating the effectiveness as well as necessity of the modeling work in Sections 3 and 4.}
\label{fig_10}
\end{figure}
\begin{figure}[!t]
\centering
\includegraphics[width=3.4in]{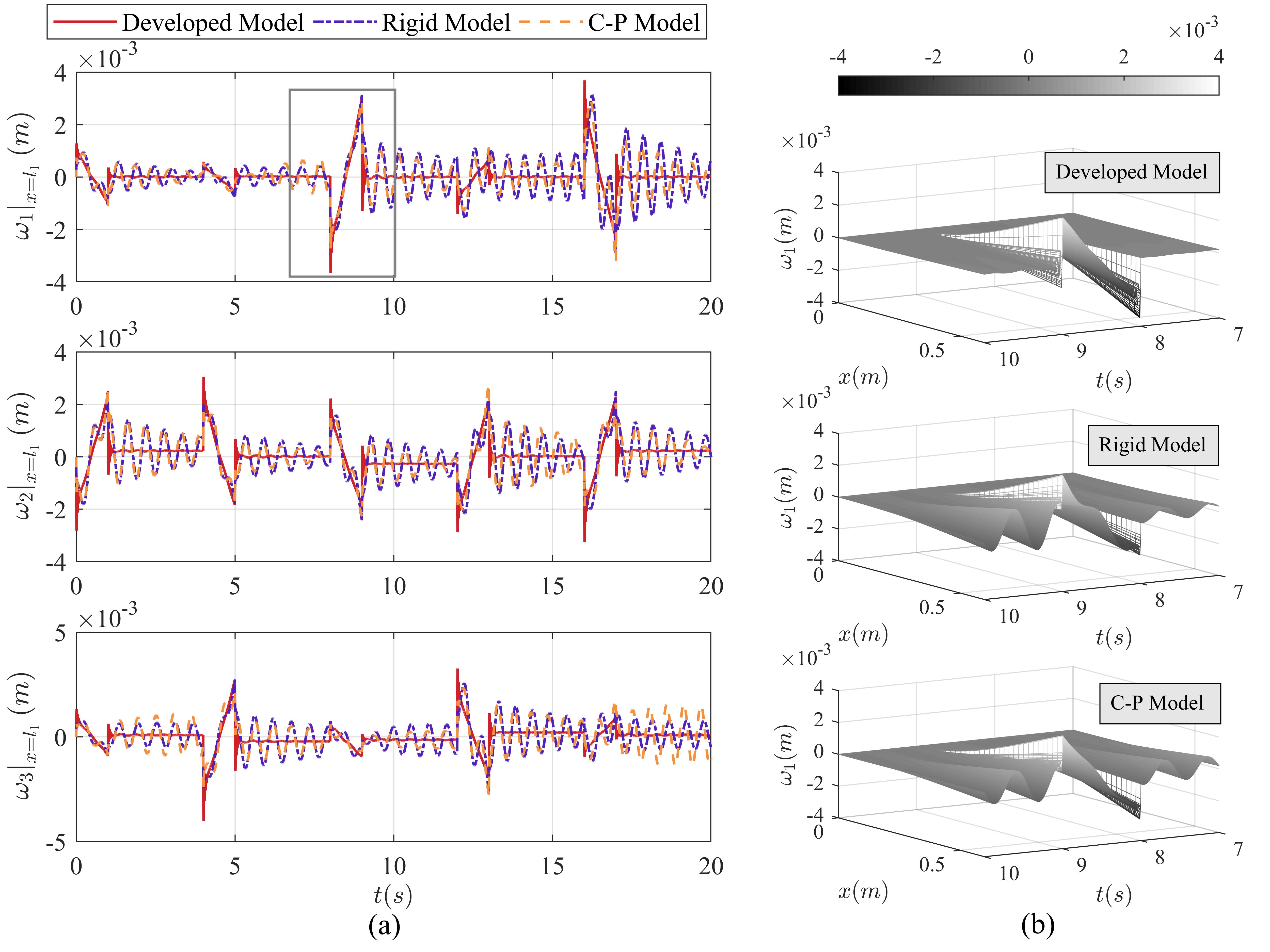}
\caption{The comparison of the deformation of the flexible links under the proposed controllers with three different dynamic models in (a) the end-point deformations of three flexible links and (b) the deformation of the entire first link corresponding to the grey box in (a).}
\label{fig_11}
\end{figure}

\subsection{Performance comparison of the proposed controller with different model compensations}
In order to show the effectiveness of the developed rigid-flexible coupling dynamic model of the 3-\underline{R}RR PMFAL, two other different dynamic models, i.e., the multi-rigid-body model (termed as “rigid model”) and the rigid-flexible coupling dynamic model with clamped-pinned boundary condition (termed as “C-P model”), are introduced and calculated in this section.

The trajectory tracking and the deformation of the flexible links of the 3-\underline{R}RR PMFAL under the proposed controllers designed with three different dynamic models (rigid model, C-P model and developed model) are shown in Fig. \ref{fig_10} and Fig. \ref{fig_11}, respectively. The results in Fig. \ref{fig_10} show that the proposed controller with three different dynamic models all provide promising trajectory tracking performance during moving phases, while the rigid model and the C-P model fail to compensate well for the system dynamics so that the residual oscillation during stabilization phases has the same magnitude as the joint-based PD control. The results in Fig. \ref{fig_11} show that the deformation of the flexible links under the proposed controller with the developed model have been dropped rapidly to around zero during stabilization phases, while the simulation results of the controller with the other two models are still oscillatory. Overall, the comparison results prove that the reliability of the developed rigid-flexible coupling dynamic model as well as the effectiveness and necessity of the data-driven identification of the mode shape function.

\subsection{Performance of the proposed controller with different update frequencies of the neural-network-based state observer}

\begin{figure}[!t]
\centering
\includegraphics[width=3.4in]{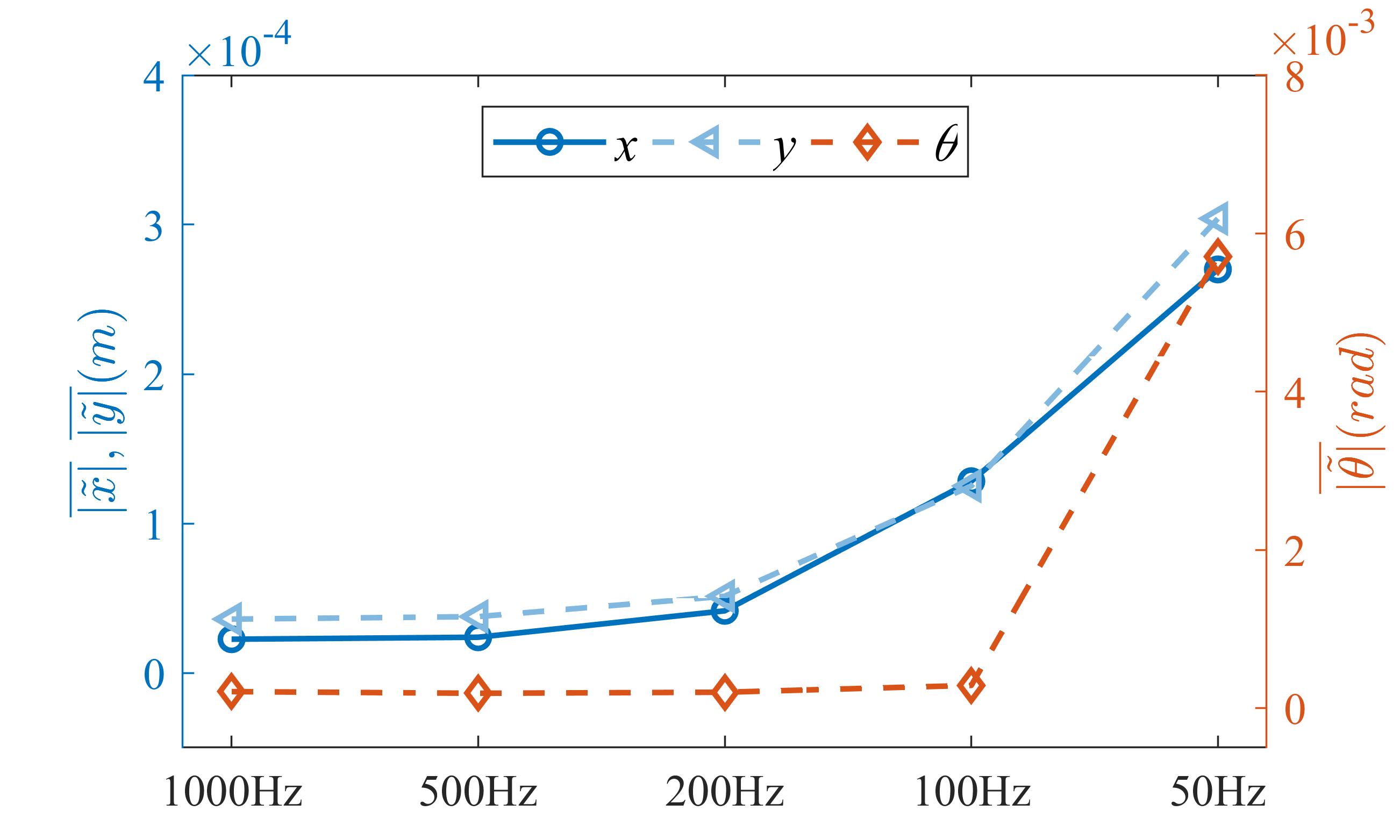}
\caption{Comparison of the averaged absolute error of trajectory tracking of the end-effector under different update frequencies of the state observer for $x$, $y$ (labeled on left vertical axis, marked in blue) and $\theta$ (labeled on right vertical axis, marked in red) during $0 - 20{\text{s}}$.}
\label{fig_12}
\end{figure}

\begin{figure}[!t]
\centering
\includegraphics[width=3.4in]{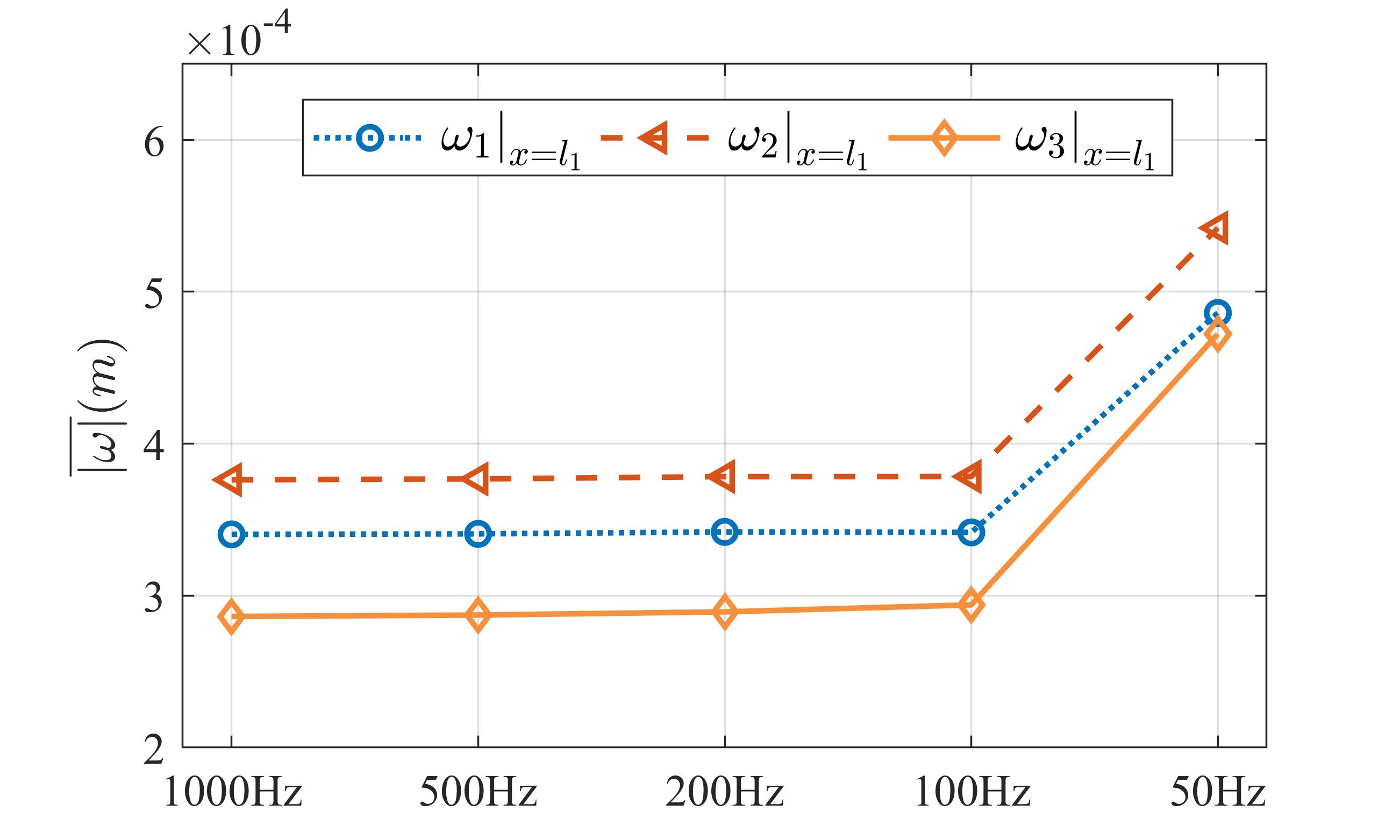}
\caption{Comparison of the end-point deformation under different update frequencies of the state observer for the three flexible links during $0 - 20{\text{s}}$.}
\label{fig_13}
\end{figure}

Herein, the neural-network-based state observer requires data of real-time deformation of flexible links. However, the feedback frequency of flexible sensing signals is usually limited in practice\cite{ref40}. Therefore, it is necessary to discuss the performance of the proposed controller with different update frequencies of the state observer. The trajectory tracking performance and the end deformation of the flexible links are shown in Figs. \ref{fig_12} and \ref{fig_13}, where in both figures the update frequency of the state observer is respectively selected as 1000Hz, 500Hz, 200Hz, 100Hz, and 50Hz. As can be seen from these figures, although the trajectory tracking error and the end-point deformation increase when decreasing the update frequency of the state observer, the proposed controller still has satisfactory performances in trajectory tracking and deformation suppression even the frequency is as low as 200Hz, implying that the control is effectively applicable for a wide range of update frequency of the state observer.

\section{Conclusions}
This study contributed to provide a promising dynamic model and design a model-based controller for the vibration suppression of the planar 3-\underline{R}RR PMFAL. Firstly, we present an inverse kinematic expression in an analytical form for the mechanism of interest, providing the basis for the symbolic derivation of the dynamic model. Secondly, to determine the mode shape function of the flexible links, the data-driven method which combines the DMD and SINDy algorithms is employed to identify the boundary condition for the flexible links in the process of AMM-based dynamic modeling. Thirdly, we construct a state observer of the end-effector by a neural network and design a model-based controller with feedback compensation on this basis. The co-simulation results show that the proposed controller which incorporates with the neural network-based state observer has promising capability to suppress the residual vibration of the flexible links, and the reliability of the developed dynamic model is revealed by comparing the control results of different models for compensation. Besides, the proposed controller also exhibits applicability for limited update frequencies of flexible sensors.

\section*{Acknowledgments}
This work was supported by the Program of the National Natural Science Foundation of China under Grant Nos. 12072237, 11872277 and the Key Program of the National Natural Science Foundation of China under Grant Nos. 11932015, 91748205.

\section*{Appendix A}
\begin{figure}[!t]
\centering
\includegraphics[width=3.4in]{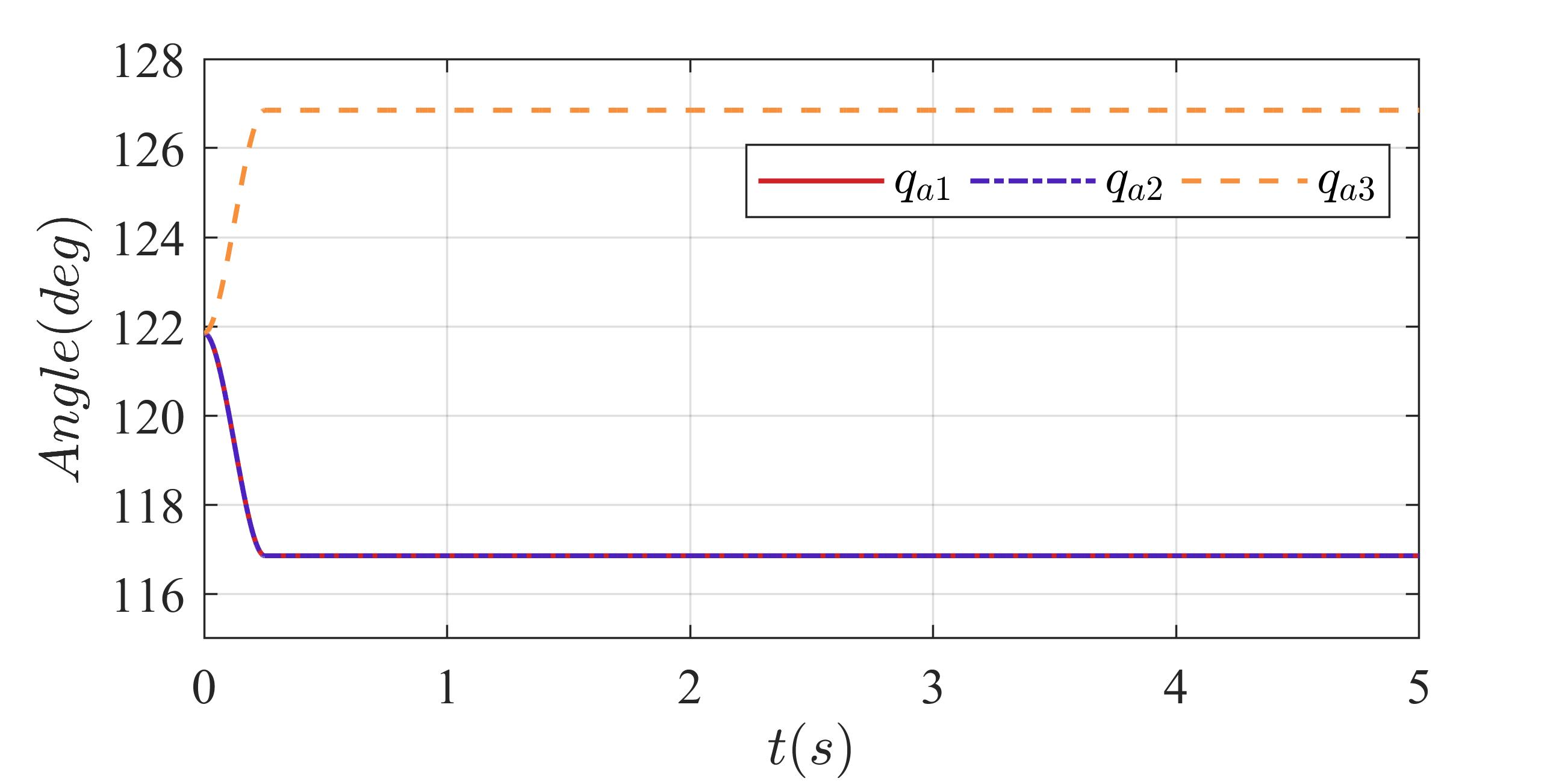}
\caption{Desired signals for active joints in the pre-simulation.}
\label{fig_14}
\end{figure}
\begin{figure}[!t]
\centering
\includegraphics[width=3.4in]{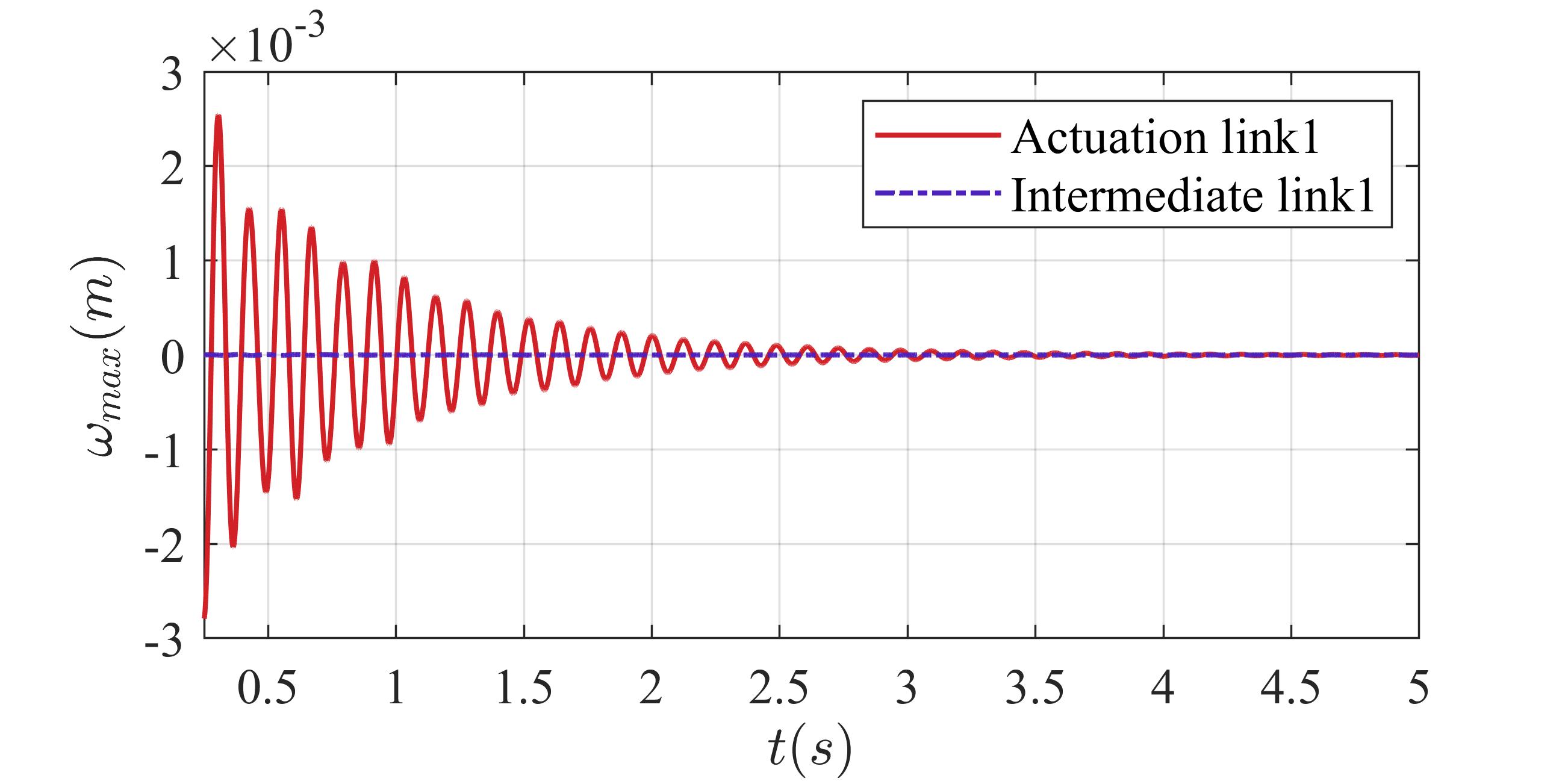}
\caption{Time history comparison between the maximum deformation points of actuation link and intermediate link.}
\label{fig_15}
\end{figure}

For the pre-simulation of the 3-\underline{R}RR PM with all the links lightweighted, the actuation links and the intermediate links are divided into finite elements by using the View/Flex module of ADAMS, and the mesh modal of each link has 9581 nodes and 4436 elements. The trajectory of the mechanism is converted into a point-to-point motion by linear interpolation in the simulation, and the desired signals for active joints are shown in Fig. \ref{fig_14}. The pre-simulation results show that the maximum deformation position of the actuation links occurs at the end point of each link, that of the intermediate links occurs at the middle point of each link, and the former is much greater than the latter (shown in Fig. \ref{fig_15}).

\section*{Appendix B}
Using the kinematic analysis in Section II, ${{\mathbf{q}}_{\text{d}}}$ and ${{\mathbf{q}}_{\text{p}}}$ can be expressed as
\begin{equation}
\label{eq_51}
\begin{gathered}
  {{\mathbf{q}}_{\text{d}}} = {{\mathbf{q}}_{\text{d}}}\left( {{{\mathbf{q}}_{\text{e}}}} \right), \hfill \\
  {{\mathbf{q}}_{\text{p}}} = {{\mathbf{q}}_{\text{p}}}\left( {{{\mathbf{q}}_{\text{e}}}} \right). \hfill \\ 
\end{gathered}
\end{equation}
Applying the kinematic constraints to Lagrange-D’ Alembert yields
\begin{equation}
\label{eq_52}
\begin{gathered}
  \left( {\frac{d}{{dt}}\left( {\frac{{\partial L}}{{\partial {\mathbf{\dot q}}_{\text{w}}^{\text{T}}}}} \right) - \frac{{\partial L}}{{\partial {\mathbf{q}}_{\text{w}}^{\text{T}}}} - {\mathbf{Q}}_{\text{w}}^{\text{T}}} \right)\delta {{\mathbf{q}}_{\text{w}}} \hfill \\
   = \left( {\frac{d}{{dt}}\left( {\frac{{\partial L}}{{\partial {\mathbf{\dot q}}_{\text{a}}^{\text{T}}}}} \right) - \frac{{\partial L}}{{\partial {\mathbf{q}}_{\text{a}}^{\text{T}}}} - {\mathbf{\tau }}_{\text{a}}^{\text{T}}} \right)\delta {{\mathbf{q}}_{\text{a}}} \hfill \\
   + \left( {\frac{d}{{dt}}\left( {\frac{{\partial L}}{{\partial {\mathbf{\dot q}}_{\text{p}}^{\text{T}}}}} \right) - \frac{{\partial L}}{{\partial {\mathbf{q}}_{\text{p}}^{\text{T}}}} - {\mathbf{\tau }}_{\text{p}}^{\text{T}}} \right)\delta {{\mathbf{q}}_{\text{p}}} \hfill \\
   + \left( {\frac{d}{{dt}}\left( {\frac{{\partial L}}{{\partial {\mathbf{\dot q}}_{\text{f}}^{\text{T}}}}} \right) - \frac{{\partial L}}{{\partial {\mathbf{q}}_{\text{f}}^{\text{T}}}} - {\mathbf{\tau }}_{\text{f}}^{\text{T}}} \right)\delta {{\mathbf{q}}_{\text{f}}} \hfill \\
   = \left( {\frac{d}{{dt}}\left( {\frac{{\partial L}}{{\partial {\mathbf{\dot q}}_{\text{d}}^{\text{T}}}}} \right) - \frac{{\partial L}}{{\partial {\mathbf{q}}_{\text{d}}^{\text{T}}}} - {{\mathbf{\tau }}^{\text{T}}}} \right)\delta {{\mathbf{q}}_{\text{d}}} \hfill \\
   + \left( {\frac{d}{{dt}}\left( {\frac{{\partial L}}{{\partial {\mathbf{\dot q}}_{\text{f}}^{\text{T}}}}} \right) - \frac{{\partial L}}{{\partial {\mathbf{q}}_{\text{f}}^{\text{T}}}} - {\mathbf{\tau }}_{\text{f}}^{\text{T}}} \right)\delta {{\mathbf{q}}_{\text{f}}} \hfill \\
   = \left( {\frac{d}{{dt}}\left( {\frac{{\partial L}}{{\partial {\mathbf{\dot q}}_{\text{d}}^{\text{T}}}}} \right) - \frac{{\partial L}}{{\partial {\mathbf{q}}_{\text{d}}^{\text{T}}}} - {{\mathbf{\tau }}^{\text{T}}}} \right)\frac{{\partial {{\mathbf{q}}_{\text{d}}}}}{{\partial {\mathbf{q}}_{\text{e}}^{\text{T}}}}\delta {{\mathbf{q}}_{\text{e}}} \hfill \\
   + \left( {\frac{d}{{dt}}\left( {\frac{{\partial L}}{{\partial {\mathbf{\dot q}}_{\text{p}}^{\text{T}}}}} \right) - \frac{{\partial L}}{{\partial {\mathbf{q}}_{\text{p}}^{\text{T}}}} - {\mathbf{\tau }}_{\text{p}}^{\text{T}}} \right)\frac{{\partial {{\mathbf{q}}_{\text{p}}}}}{{\partial {\mathbf{q}}_{\text{e}}^{\text{T}}}}\delta {{\mathbf{q}}_{\text{e}}} \hfill \\
   = 0. \hfill \\ 
\end{gathered} 
\end{equation}
Since $\delta {{\mathbf{q}}_{\text{e}}}$ is free to vary, Eq. (\ref{eq_52}) becomes
\begin{equation}
\label{eq_53}
\begin{gathered}
  \left( {\frac{d}{{dt}}\left( {\frac{{\partial L}}{{\partial {\mathbf{\dot q}}_{\text{d}}^{\text{T}}}}} \right) - \frac{{\partial L}}{{\partial {\mathbf{q}}_{\text{d}}^{\text{T}}}}} \right)\frac{{\partial {{\mathbf{q}}_{\text{d}}}}}{{\partial {\mathbf{q}}_{\text{e}}^{\text{T}}}} + \left( {\frac{d}{{dt}}\left( {\frac{{\partial L}}{{\partial {\mathbf{\dot q}}_{\text{p}}^{\text{T}}}}} \right) - \frac{{\partial L}}{{\partial {\mathbf{q}}_{\text{p}}^{\text{T}}}}} \right)\frac{{\partial {{\mathbf{q}}_{\text{p}}}}}{{\partial {\mathbf{q}}_{\text{e}}^{\text{T}}}} \hfill \\
   = \left( {\frac{d}{{dt}}\left( {\frac{{\partial L}}{{\partial {\mathbf{\dot q}}_{\text{w}}^{\text{T}}}}} \right) - \frac{{\partial L}}{{\partial {\mathbf{q}}_{\text{w}}^{\text{T}}}}} \right)\frac{{\partial {{\mathbf{q}}_{\text{w}}}}}{{\partial {\mathbf{q}}_{\text{e}}^{\text{T}}}} \hfill \\
   = {{\mathbf{\tau }}^{\text{T}}}\frac{{\partial {{\mathbf{q}}_{\text{d}}}}}{{\partial {\mathbf{q}}_{\text{e}}^{\text{T}}}} + {\mathbf{\tau }}_{\text{p}}^{\text{T}}\frac{{\partial {{\mathbf{q}}_{\text{p}}}}}{{\partial {\mathbf{q}}_{\text{e}}^{\text{T}}}}, \hfill \\ 
\end{gathered} 
\end{equation}
where ${{\mathbf{\tau }}_{\text{p}}}$ is zero if the frictions on passive joints are negligible. Then, combining Eq. (\ref{eq_22}), Eq. (\ref{eq_53}) can be rewritten as
\begin{equation}
\label{eq_54}
{{\mathbf{J}}^{\text{T}}}{\mathbf{\tau }} = {{\mathbf{S}}^{\text{T}}}{{\mathbf{Q}}_{\text{w}}}.
\end{equation}

\end{document}